\documentclass{article}

\usepackage[nonatbib,final]{neurips_2022}

\usepackage{graphicx}
\usepackage[utf8]{inputenc} 
\usepackage[T1]{fontenc}    
\usepackage{textgreek}      
\usepackage{url}            
\usepackage{booktabs}       
\usepackage{amsfonts}       
\usepackage{amsmath}        
\usepackage{nicefrac}       
\usepackage{microtype}      
\usepackage[table]{xcolor}  
\usepackage{lipsum}         
\usepackage[colorinlistoftodos]{todonotes}       
\usepackage{multirow}       
\usepackage[ruled,linesnumbered]{algorithm2e}    
\usepackage[super]{nth}     
\usepackage{arydshln}       
\usepackage{colortbl}       
\usepackage{floatrow}       
\usepackage{wrapfig}
\usepackage[export]{adjustbox}
\usepackage{xfakebold}      
\usepackage[backend=biber,style=ieee,citestyle=numeric-comp,sorting=ydnt,dashed=false]{biblatex}
\usepackage{floatrow}       

\definecolor{Gray}{HTML}{808080}
\definecolor{Red}{HTML}{FF4B8A}
\definecolor{Green}{HTML}{34a853}
\definecolor{Blue}{HTML}{008AE3}
\definecolor{Gold}{HTML}{FFAA4D}
\definecolor{SkyBlue}{HTML}{00BCE3}
\definecolor{LightPink}{HTML}{FF85BD}
\definecolor{HotPink}{HTML}{FF02FF}
\definecolor{AttnPurple}{HTML}{764ECA}
\definecolor{EmbedPink}{HTML}{FC4291}
\definecolor{MLPGreen}{HTML}{17B591}
\definecolor{HeadOrange}{HTML}{FE6F00}
\definecolor{inGold}{HTML}{FFAA4D}
\definecolor{gtGreen}{HTML}{44D62C}
\definecolor{vitcaPurple}{HTML}{EA27C2}
\definecolor{unetcaBlue}{HTML}{009ACE}

\newfloatcommand{capbtabbox}{table}[][\FBwidth]

\newcommand\gt{$\mathbf{X}$}
\newcommand\gtplain{\mathbf{X}}
\newcommand\gtplainsmall{\mathbf{x}}
\newcommand\maskedinput{\textcolor{Gold}{$\texttt{mask}(\gtplain)$}}
\newcommand\maskedinputsmall{\textcolor{Gold}{$\texttt{mask}(\gtplainsmall)$}}
\newcommand\cells{$\mathbf{Z}$}
\newcommand\cellsplain{\mathbf{Z}}

\newcommand\cellsoutput{\textcolor{SkyBlue}{$\mathbf{Z}_o$}}
\newcommand\currentoutput{\textcolor{SkyBlue}{$\mathbf{z}^t_o$}}

\newcommand\currenthidden{\textcolor{LightPink}{$\mathbf{z}^t_h$}}

\newcommand\pe{\textcolor{Gray}{${\fbseries[0.75] \gamma}$}}
\newcommand\peplain{\gamma}
\newcommand\embed{\textcolor{EmbedPink}{\texttt{embed}}}
\newcommand\localize{\textcolor{AttnPurple}{\texttt{localize}}}
\newcommand\mhsa{\textcolor{AttnPurple}{\texttt{MHSA}}}
\newcommand\sa{\texttt{SA}}
\newcommand\mlp{\textcolor{MLPGreen}{\texttt{MLP}}}
\newcommand\mlphead{\textcolor{HeadOrange}{\texttt{head}}}
\newcommand\tokens{$\mathbf{T}$}
\newcommand\pretokens{$\mathbf{T}^\prime$}
\newcommand\pool{$\mathcal{P}$}
\newcommand\poolplain{\mathcal{P}}
\newcommand\poolsize{$N_\poolplain$}

\newcommand*\rot{\rotatebox{90}}
\newcommand\diag[1]{\multicolumn{1}{l}{\rlap{\rotatebox{55}{#1}~}}}

\newcommand\psnrlabel{PSNR $\uparrow$}
\newcommand\ssimlabel{SSIM $\uparrow$}
\newcommand\lpipslabel{LPIPS $\downarrow$}
\newcommand\paramslabel{\# Params.}
\newcommand\acclabel{Acc.\ $\uparrow$}
\newcommand\fwdlabel{Fwd.\ $\downarrow$}
\newcommand\bwdlabel{Bwd.\ $\downarrow$}
\newcommand\memlabel{Mem.\ $\downarrow$}
\makeatletter
\def\adl@drawiv#1#2#3{%
        \xleaders#3{#2.5\@tempdimb #1{1}#2.5\@tempdimb}%
                #2\z@ plus1fil minus1fil\relax}
\newcommand{\cdashrule}[1]{%
  \noalign{\vskip\aboverulesep
           \global\let\@dashdrawstore\adl@draw
           \global\let\adl@draw\adl@drawiv}
  \cdashline{#1}
  \noalign{\global\let\adl@draw\@dashdrawstore
           \vskip\belowrulesep}}
\makeatother

\SetKwComment{Comment}{// }{}

\SetCommentSty{mycommfont}
\SetKwInOut{Input}{Input}
\SetKwInOut{Output}{Output}
\SetKw{and}{and}

\newcommand{\ts}{\textsuperscript}

\graphicspath{{assets/}}

\usepackage[colorlinks]{hyperref}

\usepackage[capitalize]{cleveref}
\crefname{section}{Sec.}{Secs.}
\Crefname{section}{Section}{Sections}
\Crefname{table}{Table}{Tables}
\crefname{table}{Tab.}{Tabs.}

\makeatletter
\DeclareRobustCommand\onedot{\futurelet\@let@token\@onedot}
\def\@onedot{\ifx\@let@token.\else.\null\fi\xspace}

\def\eg{\emph{e.g}\onedot, } \def\Eg{\emph{E.g}\onedot, }
\def\ie{\emph{i.e}\onedot, } 
\def\st{\text{s.t\onedot}\ }
 
\def\etc{\emph{etc}\onedot}
\def\vs{\emph{vs}\onedot}
\def\wrt{w.r.t\onedot}

\makeatother
\newcommand\Tab[1]{Tab.\ {#1}}
\newcommand\Fig[1]{Fig.\ {#1}}
\newcommand\Alg[1]{Alg.\ {#1}}
\newcommand\Sec[1]{Sec.\ {#1}}

\NewDocumentCommand{\fbseries}{ o }{%
  \IfValueTF{#1}
    {\unskip\setBold[#1]}%
    {\setBold[0.3]}
  \aftergroup\unsetBold\aftergroup
  \xspace
}

\begin{document}

\title{Attention-based Neural Cellular Automata}

\author{%
  \href{https://mtesfaldet.net}{\textcolor{LightPink}{Mattie Tesfaldet}}\\
  McGill University, Mila\\
  \And
  \href{https://www.cim.mcgill.ca/~derek/}{\textcolor{Gold}{Derek Nowrouzezahrai}}\\
  McGill University, Mila\\
  \And
  \href{https://sites.google.com/view/christopher-pal}{\textcolor{SkyBlue}{Christopher Pal}}\thanks{Canada CIFAR AI Chair}\\
  Polytechnique Montréal, Mila\\
}


\maketitle

\begin{abstract}
   Recent extensions of Cellular Automata (CA) have incorporated key ideas from modern deep learning, dramatically extending their capabilities and catalyzing a new family of Neural Cellular Automata (NCA) techniques. Inspired by Transformer-based architectures, our work presents a new class of \textit{attention-based} NCAs formed using a spatially localized---yet globally organized---self-attention scheme. We introduce an instance of this class named \emph{Vision Transformer Cellular Automata (ViTCA)}. We present quantitative and qualitative results on denoising autoencoding across six benchmark datasets, comparing ViTCA to a U-Net, a U-Net-based CA baseline (UNetCA), and a Vision Transformer (ViT). When comparing across architectures configured to similar parameter complexity, ViTCA architectures yield superior performance across all benchmarks and for nearly every evaluation metric. We present an ablation study on various architectural configurations of ViTCA, an analysis of its effect on cell states, and an investigation on its inductive biases. Finally, we examine its learned representations via linear probes on its converged cell state hidden representations, yielding, on average, superior results when compared to our U-Net, ViT, and UNetCA baselines.
\end{abstract}

\section{Introduction}
\vspace{-.5\baselineskip}
\label{sec:introduction}

\begin{wrapfigure}{r}[0pt]{0.5\textwidth}
  \vspace{-\baselineskip}
  \centerline{\includegraphics[width=\textwidth]{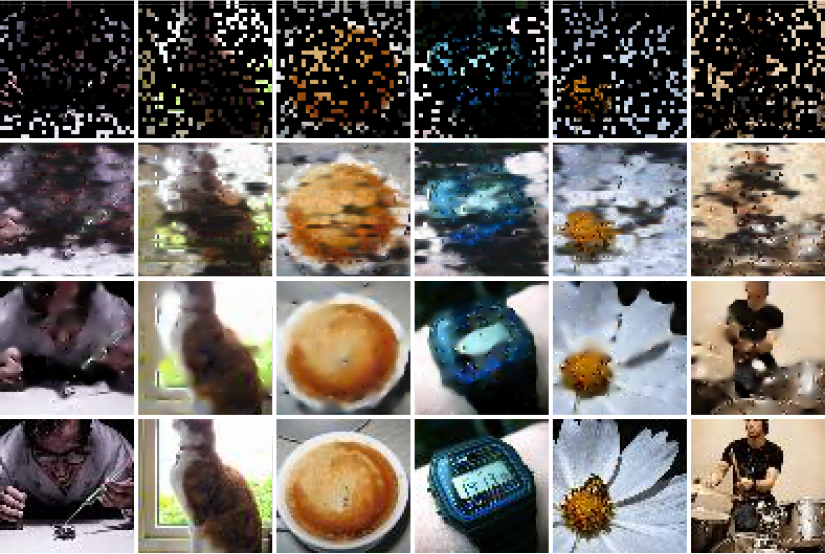}}
  \caption{\textbf{ViT} \vs\ \textbf{ViTCA} for denoising Tiny ImageNet \cite{russakovsky2015imagenet} validation set images with 2$\times$2 pixel masks covering $75\%$ of the image. \emph{Top-to-bottom}: noisy input, ViT, ViTCA, and ground truth.\vspace{-\baselineskip}} 
  \label{fig:vitca_vs_vit}
\end{wrapfigure} Recent developments at the intersection of two foundational ideas---Artificial Neural Networks (ANNs) and Cellular Automata (CA)---have led to new approaches for constructing Neural Cellular Automata (NCA). These advances have integrated ideas such as variational inference \cite{palm2022variational}, U-Nets \cite{zhang2020learning}, and Graph Neural Networks (GNNs) \cite{grattarola2021learning} with promising results on problems ranging from image synthesis \cite{palm2022variational,niklasson2021self-organising,mordvintsev2021mu} to Reinforcement Learning (RL) \cite{najarro2022hypernca,variengien2021towards}. Transformers are another significant development in deep learning \cite{vaswani2017attention}, but, until now, have not been examined under an NCA setting.

Vision Transformers (ViTs) \cite{dosovitskiy2020image} have emerged as a competitive alternative to Convolutional Neural Network (CNN) \cite{lecun1998gradient} architectures for computer vision, such as Residual Networks (ResNets) \cite{he2016deep}. ViTs leverage the self-attention mechanisms of original Transformers \cite{vaswani2017attention}, which have emerged as the dominant approach for sequence modelling in recent years. Our work combines foundational ideas from Transformers and ViTs, leading to a new class of NCAs: \textbf{Vision Transformer Cellular Automata (ViTCA)}. 

An effective and ubiquitous Transformer-based learning technique for Natural Language Processing (NLP) pre-training is the unsupervised task of Masked Language Modelling (MLM), popularized by the BERT language model \cite{devlin-etal-2019-bert}. The success of MLM-based techniques has similarly inspired recent work re-examining the classical formulation of Denoising Autoencoders (DAEs) \cite{vincent2010stacked}, but for ViTs \cite{bao2022beit,dosovitskiy2020image,chen2020generative}, introducing tasks such as Masked Image Encoding \cite{he2021masked} and Masked Feature Prediction \cite{wei2021masked} for image and video modelling, respectively. This simple yet highly-scalable strategy of masked-based unsupervised pre-training has yielded promising transfer learning results on vision-based downstream tasks such as object detection and segmentation, image classification, and action detection, even outperforming supervised pre-training~\cite{he2021masked,wei2021masked}. We examine training methodologies for ViTCA within a DAE setting and perform extensive controlled experiments benchmarking these formulations against modern state of the art architectures, with favourable outcomes, \eg \Fig{\ref{fig:vitca_vs_vit}}.

\begin{figure}[t]
  \centerline{\includegraphics[scale=0.9]{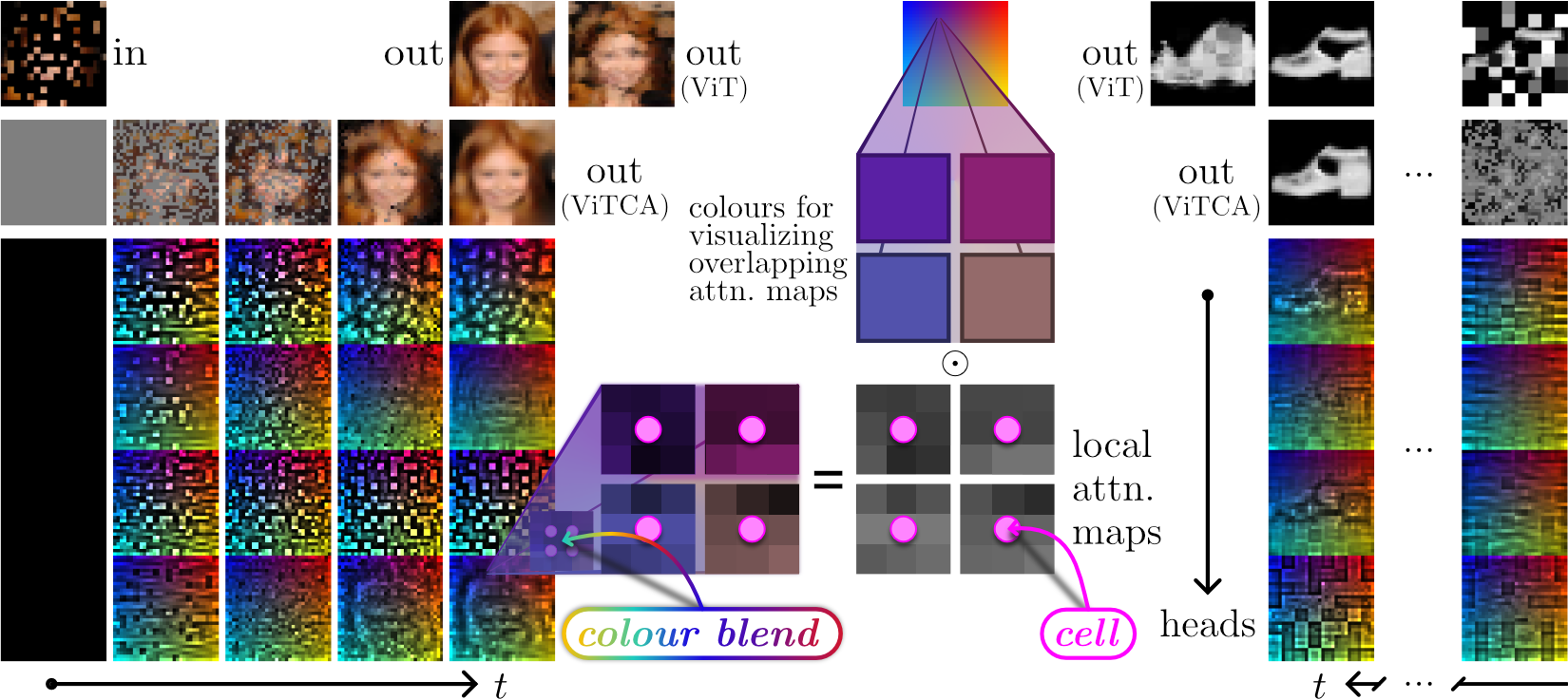}}
  \caption{\textbf{Global self-organization manifested within localized self-attention}. Despite operating in spatially local neighbourhoods about a cell, over time the localized (multi-head) self-attention in ViTCA experiences a \emph{global} self-organization admitted by its NCA nature. This circumvents the quadratic complexity of explicit global self-attention (\wrt input size) with a linear amortization over time (recurrent CA iterations), enabling effective per-pixel dense processing. \emph{Middle}: visualizing local attention maps about each cell as colour-coded ``splats'' blended together in overlapping regions, producing a ``splat map''~\cite{crawfis1993texture}. \emph{Left, right}: ViTCA iterations on a cell grid, updated from a seed state to a converged state, given a noisy input image to denoise. For each head of the cells' local attention maps, there is global agreement on the types of features to attend to (\eg foreground contours, noise, background). Enveloping ViT by the NCA paradigm dramatically improves its output fidelity.\vspace{-\baselineskip}}
  \label{fig:teaser}
\end{figure}

Our contributions are as follows: \textit{first}---to the best of our knowledge---our work is the first to extend NCA methodologies with key Transformer mechanisms, \ie self-attention and positional encoding (and embedding), with the beneficial side-effect of circumventing the quadratic complexity of self-attention; \textit{second}, our ViTCA formulation allows for lower model complexity (by limiting ViT depth) while retaining expressivity through CA iterations on a controlled state---all with the same encoder weights. This yields a demonstrably more parameter-efficient \cite{mordvintsev2021mu} ViT-based model. Importantly, ViTCA mitigates the problems associated with the explicit tuning of ViT depth originally needed to improve performance (\ie we use a depth of 1). With ViTCA, we simply iterate until cell state convergence. Since ViT (and by extension, ViTCA) employs Layer Normalization (LN) \cite{ba2016layer} at each stage of its processing, it is a fairly contractive model capable of fixed-point convergence guarantees~\cite{bai2019deep}.

In relation to our first contribution, ViTCA respects CA requirements, most importantly that computations remain localized about a cell and its neighbourhood. As such, we modify the global self-attention mechanisms of a ViT to respect this locality requirement (\Fig{\ref{fig:teaser}}). Localized self-attention is not a new idea \cite{chen2022regionvit,liu2021swin,chu2021twins,zhang2021multi}; however, because cells contain state information that depends on its previous state, over CA iterations the effective receptive field of ViTCA's localized self-attention grows increasingly larger until eventually incorporating information implicitly across all cells. Thus, admitting global propagation of information from spatially localized self-attention. Moreover, due to the self-organizing nature of NCAs, self-organization also manifests itself within the localized self-attention, resulting in a globally agreed-upon arrangement of local self-attention. Thus, circumventing the quadratic complexity of explicit global self-attention (\wrt the input size) through a linear amortization over time, and increasing the feasibility of per-pixel dense processing (as we demonstrate). This globally consistent and complex behaviour, which arises from strictly local interactions, is a unique feature of NCAs and confers performance benefits which we observe both qualitatively and quantitatively when comparing ViT and ViTCA for denoising autoencoding. 

\begin{figure}[t]
  \centerline{\includegraphics[scale=1]{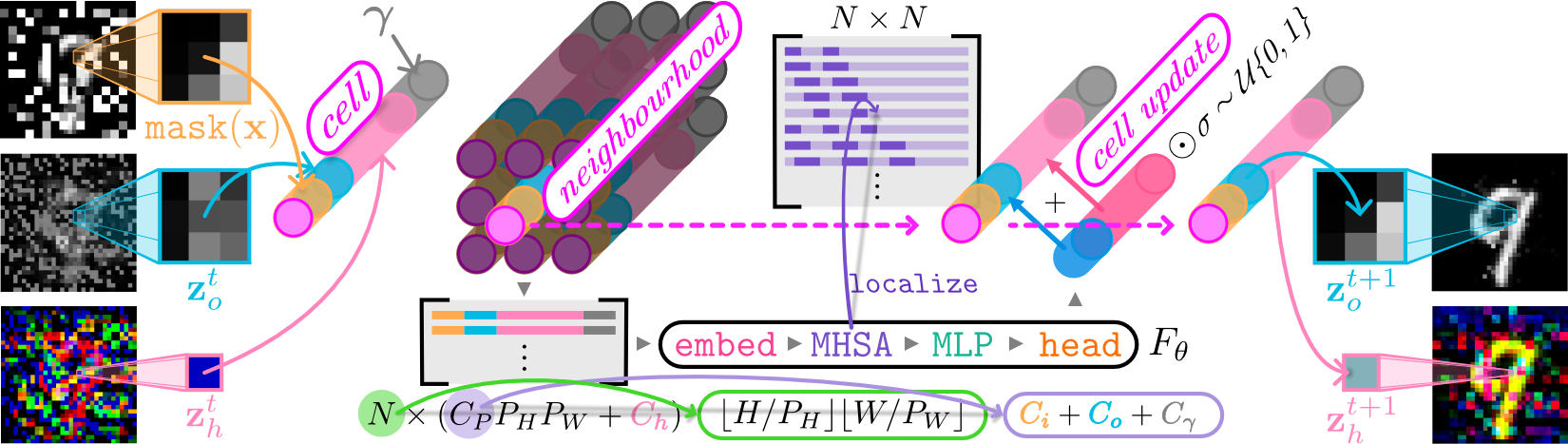}}
  \caption{\textbf{Computational overview}. NCAs use a stateful lattice of \emph{cells}, each storing information along channels, to promote desired behaviour over the course of an evolutionary cycle. Starting from an initial seed, each cell state evolves at discrete time steps according to a homogeneous, learned \emph{update rule} applied either synchronously or asynchronously ($\sigma$). This update depends on the current cell state and that of its \emph{neighbours} (pictured is the \emph{Moore neighbourhood}~\cite{weissteinmoore}). In \textbf{ViTCA}, each cell is represented as a vector where the first $C_iP_HP_W$ channels contain a $P_H \times P_W$ noisy input image patch (\maskedinputsmall), the next $C_oP_HP_W$ channels contain the current output patch (\currentoutput), the following $C_h$ channels contain undefined data hidden from the loss that can be used to encode additional information (\currenthidden), and (optionally) the remaining $C_{\peplain}P_HP_W$ channels contain positional information (\pe). The update rule ($F_\theta$) is a modified ViT \cite{dosovitskiy2020image} whose self-attention mechanism is locally constrained to each cell's neighbourhood (\localize).\vspace{-\baselineskip}}
  \label{fig:overview}
\end{figure}

\vspace{-.5\baselineskip}
\section{Background and related work}
\vspace{-.5\baselineskip}
\label{sec:related_work}

\paragraph{Neural Cellular Automata.}

Cellular Automata are algorithmic processes motivated by the biological behaviours of cellular growth and, as such, are capable of producing complex emergent (global) dynamics from the iterative application of comparatively simple (localized) rules~\cite{von1966theory}. \emph{Neural} Cellular Automata present a more general CA formulation, where the evolving cell states are represented as (typically low-dimensional) vectors and the update rule dictating their evolution is a differentiable function whose parameters are learned through backpropagation from a loss, rather than a handcrafted set of rules~\cite{mordvintsev2020growing,gilpin2019cellular,wulff1992learning}.
Neural net-based formulations of CAs in the NeurIPS community can be traced back to the early work of \cite{wulff1992learning}, where only small and simple models were examined. Recent formulations of NCAs have shown that when leveraging the power of deep learning techniques enabled by advances in hardware capabilities---namely highly-parallelizable differentiable operations implemented on GPUs---NCAs can be tuned to learn surprisingly complex desired behaviour, such as semantic segmentation \cite{sandler2020image}; common RL tasks such as cart-pole balancing \cite{variengien2021towards}, 3D locomotion \cite{najarro2022hypernca}, and Atari game playing \cite{najarro2022hypernca}; and image synthesis~\cite{palm2022variational,niklasson2021self-organising,mordvintsev2021mu}. Although these recent formulations rely on familiar compositions of convolutions and non-linear functions, it is important to highlight that NCAs are fundamentally not equivalent to ``very-deep'' CNNs (\vs~\cite{gilpin2019cellular}), or any other feedforward architecture (\eg ResNets \cite{he2016deep}), particularly, in the same way that a Recurrent Neural Network (RNN) is not equivalent: CNNs and other feedforward architectures induce a directed \textit{acyclic} computation graph (\ie a finite impulse response), whereas NCAs (and RNNs) induce a directed \textit{cyclic} computation graph (\ie an infinite impulse response), where stateful data can additionally be manipulated using (learned) feedback loops and/or time-delayed controls. As such, NCAs can be viewed as a type of RNN, and both (N)CAs and RNNs are known to be Turing complete~\cite{christen2021automatic,cook2004universality,siegelmann1995computational,wulff1992learning}.\footnote{In the case of (N)CAs, a Turing complete example is the \emph{Rule 110} elementary CA \cite{christen2021automatic,cook2004universality}}

\paragraph{Vision Transformers.}
Vision Transformers \cite{dosovitskiy2020image} are an adaptation of Transformers \cite{vaswani2017attention} to vision-based tasks like image classification. In contrast to networks built from convolutional layers, ViTs rely on \emph{self-attention} mechanisms operating on tokenized inputs. Specifically, input images are divided into non-overlapping patches, then fed to a Transformer after undergoing a linear patch projection with an embedding matrix. While ViTs provide competitive image classification performance, the quadratic computational scaling of global self-attention limits their applicability in high-dimensional domains, \eg per-pixel dense processing. Recent developments have attempted to alleviate such efficiency limitations \cite{ali2021xcit,hudson2021generative,arnab2021vivit,fan2021multiscale}, one notable example being Perceiver IO \cite{jaegle2021perceiver,yifan2021input} with its use of cross-attention. We refer interested readers to a comprehensive survey on ViTs~\cite{khan2021transformers}.

\vspace{-.5\baselineskip}
\section{Vision Transformer Cellular Automata (ViTCA)}
\vspace{-.5\baselineskip}
\label{sec:vitca}

Building upon NCAs and ViTs, we propose a new class of \emph{attention-based} NCAs formed using a spatially localized---yet globally organized---self-attention scheme. We detail an instance of this class, ViTCA, by first reviewing its backbone ViT architecture before describing the ``pool sampling''-based training process for the ViTCA update rule (see overview in \Fig{\ref{fig:overview}}).

\paragraph{Input tokenization.}

ViT starts by dividing a $C_i\!\times\!H\!\times\!W$ input image \gt\ into $N$ non-overlapping $P_H\!\times\!P_W$ patches ($16\!\times\!16$ in the original work~\cite{dosovitskiy2020image}), followed by a linear projection of the flattened image patches with an embedding matrix $\mathbf{E} \in \mathbb{R}^{L \times d}$ (\Fig{\ref{fig:overview}} \embed), where $L\!=\!C_iP_HP_W$, to produce initial tokens \pretokens\ $\!\in\!\mathbb{R}^{N \times d}$. Next, a handcrafted positional encoding \cite{vaswani2017attention} or learned positional embedding \pe\ $\in\!\mathbb{R}^{N \times d}$ \cite{dosovitskiy2020image} is added to tokens to encode positional information and break permutation invariance. Finally, a learnable class token is appended to the token sequence, resulting with \tokens\ $\!\in\!\mathbb{R}^{(N+1) \times d}$. For the purposes of our task, we omit this token in all ViT-based models. In ViTCA, the input to the embedding is a flattened cell grid \cells\ $\!\in\!\mathbb{R}^{N \times L}$ where $L\!=\!C_PP_HP_W+C_h$, $C_P\!=\!C_i+C_o+C_{\peplain}$,\, $C_h$ is the cell hidden size,\, $C_o$ is the number of output image channels (one or three for grayscale or RGB), and $C_{\peplain}$ is the positional encoding size when positional encoding is (optionally) concatenated to each cell rather than added to the tokens \cite{mildenhall2020nerf}.

\paragraph{Multi-head self-attention (MHSA).}

Given a sequence of tokens \tokens, self-attention estimates the relevance of one token to all others (\eg which image patches are likely to appear together in an image) and aggregates this global information to update each token. This encodes each token in terms of global contextual information, and does so using three learned weight matrices: $\mathbf{W}_Q\!\in\!\mathbb{R}^{d \times d}$, $\mathbf{W}_K\!\in\!\mathbb{R}^{d \times d}$, and $\mathbf{W}_V\!\in\!\mathbb{R}^{d \times d}$. \tokens\ is projected onto these weight matrices to obtain Queries $\mathbf{Q}\!=\!$ \tokens$\mathbf{W}_Q$, Keys $\mathbf{K}\!=\!$ \tokens$\mathbf{W}_K$, and Values $\mathbf{V}\!=\!$ \tokens$\mathbf{W}_V$. The self-attention layer output $\sa\!\in\!\mathbb{R}^{N \times d}$ is:

\begin{tabular}{m{5.25in}}
  \centering
  \vspace{-1.4\baselineskip}
  \begin{equation}
    \hspace{0.08in}
    \sa = \texttt{softmax}{\left({\mathbf{QK}^T}\big/{\sqrt{d}}\right)\mathbf{V}}\ .
    \label{eq:sa}
  \end{equation}
  \vspace{-1.4\baselineskip}
\end{tabular}

\emph{Multi-head} self-attention employs many sets of weight matrices, $\{\mathbf{W}_{Q_i},$ $\mathbf{W}_{K_i},$ $\mathbf{W}_{V_i}\!\in\!\mathbb{R}^{d \times (d/h)}\!\mid$  $i\!=\!0,...,(h-1)\}$. The outputs of $h$ self-attention \emph{heads} are concatenated into $(\sa_0,$ $...,$ $\sa_{h-1})\!\in\!\mathbb{R}^{N \times d}$ and projected onto a weight matrix $\mathbf{W}\!\in\!\mathbb{R}^{d \times d}$ to produce $\mhsa\!\in\!\mathbb{R}^{N \times d}$. Self-attention explicitly models global interactions and is more flexible than grid-based operators (\eg convolutions) \cite{perez2019turing,cordonnier2019relationship}, but its quadratic cost in time and memory limits its applicability to high resolution images.

\paragraph{Spatially localizing self-attention.}

The global nature of self-attention directly conflicts with the spatial locality constraint of CAs; in response, we limit the connectivity structure of the attention operation to each cell's neighbourhood. This can be accomplished by either masking each head's attention matrix ($\mathbf{A}\!=\!\texttt{softmax}(\cdots) \in \mathbb{R}^{N \times N}$ in Eq.\ \ref{eq:sa}) with a banded matrix representing local connectivity (\eg \Fig{\ref{fig:overview}} \localize), or more efficiently,

\begin{tabular}{m{3.6in} c | m{1.25in}}
  \centering
  \vspace{-1.4\baselineskip}
  \begin{equation}
    \hspace{-1em}\mathbf{A}^{\!\star} = \texttt{softmax}{\left({\mathbf{A}^\prime}\big/{\sqrt{d}}\right)} \quad
    \st \left(\mathbf{A}^\prime\right)_{ij} = \sum_{l}(\mathbf{Q})_{il}(\mathbf{K})_{jl}
    \label{eq:localize_attn}
  \end{equation}
  \vspace{-1.4\baselineskip}
  & \text{\hspace{0.5em}} &
  \vspace{-1.4\baselineskip}
  \begin{equation}
    \sa^{\!\star} = \mathbf{A}^{\!\star}\mathbf{V}
    \label{eq:localized_attn}
  \end{equation}
  \vspace{-1.4\baselineskip}
\end{tabular}

with $\left( \mathbf{V} \right)_{jl}$ where $i\!=\!\{0,...,(N\!-\!1)\}$,  $j\!=\!\{(i+n_w+n_h),...,i,...,(i-n_w-n_h)\}$, and $l\!=\!\{0,...,(d-1)\}$, and with  $n_w\!=\!\{-\lfloor{N_W/2}\rfloor,...,0,...,\lfloor{N_W/2}\rfloor\}$ and  $n_h\!=\!\{-W\lfloor{N_H/2}\rfloor,...,0,...,W\lfloor{N_H/2}\rfloor\}$. Here, we assume top-left-to-bottom-right input flattening.
Instead of explicitly computing the global self-attention matrix $\mathbf{A}\!\in\!\mathbb{R}^{N \times N}$ then masking it, this approach circumvents the $\mathcal{O}(N^2d)$ computation in favour of an $\mathcal{O}(N\!M\!d)$ alternative that indexes the necessary rows and columns \emph{during} self-attention. The result is a localized self-attention matrix $\mathbf{A}^{\!\star}\!\in\!\mathbb{R}^{N \times M}$, where $M\!=\!N_HN_W\!\ll\!N$. As we show in our experiments, ViTCA is still capable of global self-attention despite its localization, by leveraging stored state information across cells and their global self-organization during CA iterations (\Fig{\ref{fig:teaser}}).

Following \mhsa\ is a multilayer perceptron (\Fig{\ref{fig:overview}} \mlp) with two layers and a GELU non-linearity. We apply Layer Normalization (LN) \cite{ba2016layer} before \mhsa\ and \mlp, and residual connections afterwards, forming a single encoding block. We use an MLP head (\Fig{\ref{fig:overview}} \mlphead) to decode to a desired output, with LN applied to its input, finalizing the ViTCA update rule $F_\theta$. In our experiments, ViT's \mlphead\ decodes directly into an image output whereas ViTCA decodes into update vectors added to cells.

\vspace{-.5\baselineskip}
\subsection{Update rule training procedure}
\vspace{-.5\baselineskip}
\label{subsec:update_rule}

To train the ViTCA update rule, we follow a ``pool sampling''-based training process \cite{palm2022variational,mordvintsev2020growing} along with a curriculum-based masking/noise schedule when corrupting inputs. During odd training iterations, we uniformly initialize a minibatch of cells \cells\ $\!=\!(\cellsplain_1,...,\cellsplain_b)$ with constant values (0.5 for output channels, 0 for hidden---see Appendix \ref{subsec:extended_ablation} for alternatives), then inject the masked input \maskedinput\ (see \Sec{\ref{subsec:denoising_autoencoding}}). After input injection, we asynchronously update cells ($\sigma\!=\!50\%$ update rate) using $F_\theta$ for $T\!\sim\!\mathcal{U}\{8,32\}$ recurrent iterations. We retrieve output \cellsoutput\ from the cell grid and apply an $L_1$ loss against the ground truth \gt. We also apply overflow losses to penalize cell output values outside of [0,1] and cell hidden values outside of [-1,1]. We use $L_2$ normalization on the gradient of each parameter in $\theta$. After backpropagation, we append the updated cells and their ground truths to a pool \pool\ which we then shuffle and truncate up to the first \poolsize\ elements. During even training iterations, we retrieve a minibatch of cells and their ground truths from \pool\ and process them as above. This encourages $F_\theta$ to guide cells towards a stable fixed-point. \Alg{\ref{alg:vitca_training}} in Appendix \ref{sec:appendix} details this process.

\vspace{-.5\baselineskip}
\section{Experiments}
\vspace{-.5\baselineskip}
\label{sec:experiments}

Here we examine ViTCA through extensive experiments. We begin with experiments for denoising autoencoding, then an ablation study followed by various qualitative analyses, before concluding with linear probing experiments on the learned representations for MNIST \cite{deng2012mnist}, FashionMNIST \cite{xiao2017fashion}, and CIFAR10 \cite{krizhevsky2009learning}. We provide an extension to our experiments in Appendix \ref{sec:appendix}.

\paragraph{Baseline models and variants.}
%
Since we are performing pixel level reconstructions, we create a ViT baseline in which the class token has been removed. This applies identically for ViTCA. Unless otherwise stated, for our ViT and ViTCA models we use a patch size of $1\!\times\!1$ ($P_H\!=\!P_W\!=\!1$), and only a single encoding block with $h\!=\!4$ \mhsa\ heads, \embed\ size $d\!=\!128$, and \mlp\ size of $128$. For ViTCA, we choose $N_H\!=\!3$ and $N_W\!=\!3$ (\ie the \emph{Moore neighbourhood}~\cite{weissteinmoore}). We also compare with a U-Net baseline similar to the original formulation \cite{ronneberger2015u}, but based on the specific architecture from \cite{lehtinen2018noise2noise}. Since most of our datasets consist of $32\!\times\!32$ (resampled) images, we only have two downsampling steps as opposed to five. We implement a U-Net-based CA (UNetCA) baseline consisting of a modified version of our U-Net with 48 initial output feature maps as opposed to 24 and with all convolutions except the first changed to $1\!\times\!1$ to respect typical NCA restrictions~\cite{palm2022variational,mordvintsev2020growing}.

\vspace{-.5\baselineskip}
\subsection{Denoising autoencoding}
\vspace{-.5\baselineskip}
\label{subsec:denoising_autoencoding}

We compare between our baseline models and a number of ViTCA variants in the context of denoising autoencoding. We present test set results across six benchmark datasets: a land cover classification dataset intended for representation learning (LandCoverRep) \cite{yeh2021sustainbench}, MNIST, CelebA \cite{liu2015faceattributes}, FashionMNIST, CIFAR10, and Tiny ImageNet (a subset of ImageNet \cite{russakovsky2015imagenet}). All datasets consist of $32\!\times\!32$ resampled images except Tiny ImageNet, which is at $64\!\times\!64$ resolution. During testing, we use all masking combinations, chosen in a fixed order, and we update cells using a fixed number of iterations ($T\!=\!64$). See \Tab{\ref{tab:denoising_results}} for quantitative results.

Briefly mentioned in \Sec{\ref{subsec:update_rule}}, we employ a masking strategy inspired by Curriculum Learning (CL) \cite{wang2021survey,bengio2009curriculum} to ease training. This schedule follows a geometric progression of difficulty---tied to training iterations---maxing out at 10K training iterations. Specifically, masking starts at covering 25\% of the input with $1\!\times\!1$ patches of noise (dropout for RGB inputs, Gaussian for grayscale), then at each shift in difficulty, new masking configurations are added to the list of available masking configurations in the following order: $(2^0\!\times\!2^0, 50\%), (2^0\!\times\!2^0, 75\%), (2^1\!\times\!2^1, 25\%), (2^1\!\times\!2^1, 50\%), (2^1\!\times\!2^1, 75\%), ..., (2^2\!\times\!2^2, 75\%)$. Masking configurations are randomly chosen from this list.

We initialize weights/parameters using He initialization \cite{he2015delving}, except for the final layer of CA-based models, which are initialized to zero \cite{mordvintsev2020growing}. Unless otherwise stated, we train for $I\!=\!100$K iterations, use a minibatch size $b\!=\!32$, AdamW optimizer \cite{loshchilov2019decoupled}, learning rate $\eta\!=\!10^{-3}$ with a cosine annealing schedule \cite{loshchilov2017sgdr}, pool size \poolsize\ $\!=\!1024$, and cell hidden channel size $C_h\!=\!32$. In the case of Tiny ImageNet, $b\!=\!8$ to accommodate training on a single GPU (48GB Quadro RTX 8000). Training typically lasts a day at most, depending on the model. Due to the recurrent iterations required per training step, CA-based models take the longest to train. To alleviate memory limitations for some of our experiments, we use gradient checkpointing \cite{chen2016training} during CA iterations at the cost of backpropagation duration and slight variations in gradients due to its effect on round-off propagation. We also experiment with a cell fusion and mitosis scheme as an alternative. See Appendix \ref{sec:appendix} for details on runtime performance, gradient checkpointing, and fusion and mitosis.

\begin{table}[t!]
  \caption{Comparing denoising autoencoding results between baselines and ViTCA variants. ViTCA variants include: 32 (32 heads), 16 (16 heads), i (inverted bottleneck), xy (xy-coordinate positional encoding). Boldface and underlined values denote the best and second best results. Metrics include Peak Signal-to-Noise Ratio (PSNR; dB), Structural Similarity Index Measure (SSIM; values in $[0,1]$) \cite{wang2004image}, Learned Perceptual Image Patch Similarity (LPIPS; values in $[0, 1]$) \cite{zhang2018unreasonable}.\vspace{-\baselineskip}}
  \label{tab:denoising_results}
  \centering
  \small
  \setlength{\tabcolsep}{3pt}
    \begin{tabular}{@{}*{14}{l}@{}}
        & & \multicolumn{4}{c}{\textbf{LandCoverRep}} & \multicolumn{4}{c}{\textbf{CelebA}} & \multicolumn{4}{c}{\textbf{MNIST}} \\
        \cmidrule[\heavyrulewidth](lr){3-6} \cmidrule[\heavyrulewidth](lr){7-10}
        \cmidrule[\heavyrulewidth](lr){11-14}
        \addlinespace[-\aboverulesep]
        & & \diag{\psnrlabel} & \diag{\ssimlabel} & \diag{\lpipslabel} & \diag{\paramslabel} & \diag{\psnrlabel} & \diag{\ssimlabel} & \diag{\lpipslabel} & \diag{\paramslabel} & \diag{\psnrlabel} & \diag{\ssimlabel} & \diag{\lpipslabel} & \diag{\paramslabel} \\
        \toprule
        \multirow{4}{*}{\rot{\textbf{Baselines}}}
        & U-Net & 33.94 & 0.934 & \textbf{0.099} & 106.6K & 26.23 & 0.906 & 0.075 & 106.6K & 23.43 & 0.897 & 0.049 & 104.5K \\
        & ViT & 30.64 & 0.893 & 0.135 & 83.9K & 19.70 & 0.779 & 0.237 & 83.9K & 16.02 & 0.631 & 0.254 & 83.4K \\
        & UNetCA & 33.94 & \underline{0.935} & \underline{0.102} & 54.0K & 25.66 & 0.882 & 0.091 & 54.0K & 25.61 & 0.929 & 0.034 & 52.0K \\
        & ViTCA & 33.80 & 0.932 & \underline{0.102} & 92.5K & 26.53 & \underline{0.913} & \underline{0.066} & 92.5K & \underline{27.01} & 0.940 & \underline{0.028} & 91.7K \\
        \cdashrule{1-14}
        \multirow{6}{*}{\rot{\textbf{Variants}}}
        & ViTCA-32 & \underline{34.00} & \underline{0.935} & 0.103 & 92.5K & \textbf{27.01} & \textbf{0.920} & \textbf{0.060} & 92.5K & \textbf{27.68} & \textbf{0.946} & \textbf{0.026} & 91.7K \\
        & ViTCA-32xy & \textbf{34.06} & \textbf{0.936} & 0.106 & 92.8K & \underline{26.75} & 0.898 & 0.072 & 92.8K & 26.97 & \underline{0.942} & \underline{0.028} & 92.0K \\
        & ViTCA-i & 33.49 & 0.929 & 0.108 & 54.7K & 26.10 & 0.904 & 0.074 & 54.7K & 26.03 & 0.930 & 0.033 & 54.3K \\
        & ViTCA-i16 & 33.74 & 0.932 & 0.106 & 54.7K & 26.61 & 0.912 & \underline{0.066} & 54.7K & 26.42 & 0.935 & 0.031 & 54.3K \\
        & ViTCA-ixy & 33.75 & 0.933 & 0.107 & 54.8K & 26.51 & 0.894 & 0.076 & 54.8K & 25.95 & 0.933 & 0.033 & 54.4K \\
        & ViTCA-i16xy & 33.93 & \underline{0.935} & 0.108 & 54.8K & 26.68 & 0.898 & 0.074 & 54.8K & 26.28 & 0.936 & 0.031 & 54.4K \\
        \bottomrule
    \end{tabular}
    \begin{tabular}{@{}*{14}{l}@{}}
        \addlinespace[\aboverulesep]
        & & \multicolumn{4}{c}{\textbf{FashionMNIST}} & \multicolumn{4}{c}{\textbf{CIFAR10}} & \multicolumn{4}{c}{\textbf{Tiny ImageNet}} \\
        \addlinespace[-\aboverulesep]
        \cmidrule[\heavyrulewidth](lr){3-6} \cmidrule[\heavyrulewidth](lr){7-10}
        \cmidrule[\heavyrulewidth](lr){11-14}
        \addlinespace[-1pt]
        \toprule
        \multirow{4}{*}{\rot{\textbf{Baselines}}}
        & U-Net & 24.19 & 0.852 & 0.126 & 104.5K & 25.62 & 0.855 & 0.131 & 106.6K & 21.93 & 0.775 & 0.203 & 106.6K \\
        & ViT & 16.28 & 0.519 & 0.397 & 83.4K & 20.99 & 0.744 & 0.237 & 83.9K & 17.80 & 0.598 & 0.355 & 83.9K \\
        & UNetCA & 23.67 & 0.854 & 0.123 & 52.0K & 25.49 & 0.851 & 0.129 & 54.0K & 21.78 & 0.773 & 0.204 & 54.0K \\
        & ViTCA & 23.80 & 0.855 & 0.117 & 91.7K & 25.61 & 0.856 & 0.127 & 92.5K & 21.58 & 0.772 & 0.215 & 92.5K \\
        \cdashrule{1-14}
        \multirow{6}{*}{\rot{\textbf{Variants}}}
        & ViTCA-32 & \textbf{24.91} & \textbf{0.874} & \textbf{0.098} & 91.7K & \underline{26.05} & \underline{0.864} & \underline{0.122} & 92.5K & 21.94 & 0.781 & 0.202 & 92.5K \\
        & ViTCA-32xy & \underline{24.55} & \underline{0.869} & \underline{0.102} & 92.0K & \textbf{26.14} & \textbf{0.866} & \textbf{0.120} & 92.8K & \textbf{22.03} & \textbf{0.783} & \textbf{0.199} & 92.8K \\
        & ViTCA-i & 22.84 & 0.827 & 0.139 & 54.3K & 25.42 & 0.853 & 0.132 & 54.7K & 21.75 & 0.776 & 0.211 & 54.7K \\
        & ViTCA-i16 & 23.32 & 0.839 & 0.127 & 54.3K & 25.65 & 0.856 & 0.128 & 54.7K & 21.72 & 0.774 & 0.213 & 54.7K \\
        & ViTCA-ixy & 23.54 & 0.848 & 0.123 & 54.4K & 25.85 & 0.861 & 0.125 & 54.8K & 21.95 & \underline{0.782} & \underline{0.201} & 54.8K \\
        & ViTCA-i16xy & 23.59 & 0.848 & 0.121 & 54.4K & 25.98 & 0.863 & 0.123 & 54.8K & 21.99 & \underline{0.782} & \underline{0.201} & 54.8K \\
        \bottomrule
    \end{tabular}
\end{table}

Amongst baselines, ViTCA outperforms on most metrics across the majority of datasets used (10 out of 18). Exceptions include LandCoverRep, where UNetCA universally outperforms by a small margin, likely due to the texture-dominant imagery being amenable to convolutions. Notably, ViTCA strongly outperforms on MNIST. Although MNIST is a trivial dataset for common tasks such as classification, our masking/noise strategy turns it into a challenging dataset for denoising autoencoding, \eg it is difficult for even a human to classify a $32\!\times\!32$ MNIST digit 75\% corrupted by $4\!\times\!4$ patches of Gaussian noise. We hypothesize that when compared to convolutional models, ViTCA's weaker inductive biases (owed to attention \cite{yifan2021input,jaegle2021perceiver}) immediately outperform these models when there are large regions lacking useful features, \eg MNIST digits cover a small space in the canvas. This is not the case with FashionMNIST, where the content is more filled out. Between baselines and ViTCA variants, ViTCA-32 (32 heads) and 32xy (xy-coordinate positional encoding) outperform all models by large margins, demonstrating the benefits of multi-head self-attention. We also experiment with a parameter-reduced (by $\sim\!60\%$), inverted bottleneck variant where $d\!=\!64$ and \mlp\ size is 256, often with a minimal reduction in performance.
%
\vspace{-.5\baselineskip}
\subsubsection{Ablation study}
\label{subsubsec:ablation_study}
\vspace{-.5\baselineskip}

In \Tab{\ref{tab:ablation_results}} we perform an ablation study using the baseline ViTCA model above as reference on CelebA. Results are ordered in row-wise blocks, top-to-bottom. Specifically, we examine the impact of varying the cell hidden size $C_h$; the \embed\ size $d$; the number of \mhsa\ heads $h$; the depth (\# encoders), comparing both ViTCA (used throughout the table) with ViT; and in the last block we examine the impact of various methods of incorporating positional information into the model.

\begin{wraptable}[44]{r}{0.515\textwidth}
  \caption{Quantitative ablation for denoising autoencoding with ViTCA (unless otherwise stated via prefix) on CelebA \cite{liu2015faceattributes}. Boldface and underlining denote best and second best results. Italicized items denote baseline configuration settings. \ts\textdagger Trained with gradient checkpointing \cite{chen2016training}, which slightly alters round-off error during backpropagation, resulting in slight variations of results compared to training without checkpointing. See Appendix \ref{subsec:extended_ablation}.}
  \label{tab:ablation_results}
  \centering
  \small
  \setlength{\tabcolsep}{3pt}
    \begin{tabular}{@{}*{6}{l}@{}}
        & & {\psnrlabel} & {\ssimlabel} & {\lpipslabel} & {\paramslabel} \\
        \toprule
        \multirow{7}{*}{\rot{\textbf{Hidden dim}}}
        & 8 & 25.61 & 0.898 & 0.086 & 86.3K \\
        & 16 & 26.11 & 0.909 & 0.070 & 88.4K \\
        & \textit{32} & 26.53 & 0.913 & \underline{0.066} & 92.5K \\
        & 64 & 26.53 & 0.913 & \underline{0.066} & 100.7K \\
        & 128 & 26.51 & 0.912 & \underline{0.066} & 117.2K \\
        & 256 & \underline{26.77} & \underline{0.915} & \textbf{0.063} & 150.1K \\
        & 512 & \textbf{26.78} & \textbf{0.916} & \textbf{0.063} & 215.9K \\
        \midrule[\heavyrulewidth]
        \multirow{7}{*}{\rot{\textbf{Embed dim}}}
        & 8\ts\textdagger & 21.67 & 0.814 & 0.258 & 2.0K \\
        & 16\ts\textdagger & 23.22 & 0.853 & 0.183 & 4.5K \\
        & 32\ts\textdagger & 24.94 & 0.875 & 0.110 & 10.9K \\
        & 64\ts\textdagger & 25.69 & 0.898 & 0.084 & 29.9K \\
        & \textit{128}\ts\textdagger & \underline{26.05} & \underline{0.904} & \underline{0.075} & 92.5K \\
        & 256\ts\textdagger & \textbf{26.36} & \textbf{0.911} & \textbf{0.067} & 316.0K \\
        & 512\ts\textdagger & 19.93 & 0.768 & 0.274 & 1.2M \\
        \midrule[\heavyrulewidth]
        \multirow{6}{*}{\rot{\textbf{Heads}}}
        & 1 & 25.01 & 0.890 & 0.096 & 76.0K \\
        & \textit{4} & 26.53 & 0.913 & 0.066 & 92.5K \\
        & 8 & 26.77 & 0.916 & 0.062 & 92.5K \\
        & 16 & 26.78 & 0.917 & 0.062 & 92.5K \\
        & 32 & \textbf{27.01} & \textbf{0.920} & \textbf{0.060} & 92.5K \\
        & 64 & \underline{26.94} & \underline{0.919} & \underline{0.061} & 92.5K \\
        \midrule[\heavyrulewidth]
        \multirow{6}{*}{\rot{\textbf{Depth}}}
        & \emph{ViTCA--1} & \textbf{26.53} & \textbf{0.913} & \textbf{0.066} & 92.5K \\
        & ViTCA--2\ts\textdagger & \underline{10.82} & \underline{0.225} & \underline{0.771} & 175.3K \\
        & ViTCA--3\ts\textdagger & 9.70 & 0.165 & 0.793 & 258.0K \\
        \cmidrule(l){2-6}
        & \emph{ViT--1} & 19.70 & 0.779 & 0.237 & 83.9K \\
        & ViT--2\ts\textdagger & \underline{25.20} & \underline{0.900} & \underline{0.074} & 166.7K \\
        & ViT--3\ts\textdagger & \textbf{26.10} & \textbf{0.914} & \textbf{0.065} & 249.4K \\
        \midrule[\heavyrulewidth]
        \multirow{6}{*}{\rot{\textbf{PE type}}}
        & sincos5 & \underline{26.92} & \underline{0.917} & \underline{0.062} & 95.1K \\
        & sincos5xy & \textbf{27.00} & \textbf{0.919} & \textbf{0.059} & 95.3K \\
        & xy & 26.45 & 0.894 & 0.077 & 92.8K \\
        & \textit{handcrafted} & 26.53 & 0.913 & 0.066 & 92.5K \\
        & learned & 26.16 & 0.910 & 0.071 & 223.6K \\
        & none & 26.28 & 0.890 & 0.081 & 92.5K \\
        \bottomrule
        &&&&&\\ 
    \end{tabular}
    \vspace{-\baselineskip}
\end{wraptable}

Specifically, we examine the use of: (1) a xy-coordinate-based positional encoding \emph{concatenated} (``injected'') to cells, and; (2) a Transformer-based positional encoding (or embedding, if learned) \emph{added} into \embed. These two categories are subdivided into: 
(1a) sincos5---consisting of handcrafted Fourier features \cite{mildenhall2020nerf} with four doublings of a base frequency, \ie \pe\ $\!=\!(\sin{2^0\pi p},$ $\cos{2^0\pi p},$ $...,\sin{2^{J-1}\pi p},$ $\cos{2^{J-1}\pi p})\!\in\mathbb{R}^{N \times (4JP_HP_W)}$ where $J\!=\!5$ and $p$ is the pixel coordinate (normalized to [-1,1]) for each pixel the cell is situated on (one pixel since $P_H\!=\!P_W\!=\!1$); 
(1b) sincos5xy---consisting of both Fourier features and explicit xy-coordinates concatenated; 
(1c) xy---only xy-coordinates; 
(2a) handcrafted (our baseline approach)---sinusoidal encoding \pe\ $\!\in\!\mathbb{R}^{N \times d}$ similar to (1a) but following a Transformer-based approach \cite{vaswani2017attention}, and; 
(2b) learned---learned embedding \pe\ $\!\in\!\mathbb{R}^{N \times d}$ following the original ViT approach \cite{dosovitskiy2020image}. 
To further test the self-organizing capabilities of ViTCA, we also include: (3) none---no explicit positioning provided, where we let the cells localize themselves.

As shown in \Tab{\ref{tab:ablation_results}}, ViTCA benefits from an increase to most CA and Transformer-centric parameters, at the cost of computational complexity and/or an increase in parameter count. A noticeable decrease in performance is observed when \embed\ size $d\!=\!512$, most likely due to the vast increase in parameter count necessitating more training. In the original ViT, multiple encoding blocks were needed before the model could exhibit performance equivalent to their baseline CNN \cite{dosovitskiy2020image}, as verified in our ablation with our ViT. However, for ViTCA we notice an inverse relationship of the effect of Transformer depth, causing a divergence in cell state. It is not clear why this is the case, as we have observed that the LN layers and overflow losses otherwise encourage a contractive $F_\theta$. This is an investigation we leave for future work. Despite the benefits of increasing $h$, we use $h\!=\!4$ for our baseline to optimize runtime performance. Finally, we show that ViTCA does not dramatically suffer when no explicit positioning is used---in contrast to typical Transformer-based models---as cells are still able to localize themselves by relying on their stored hidden information.

\vspace{-.5\baselineskip}
\subsubsection{Cell state analysis}
\label{subsubsec:cell_state_analysis}
\vspace{-.5\baselineskip}

\begin{figure}[t]
  \centerline{\includegraphics[scale=1]{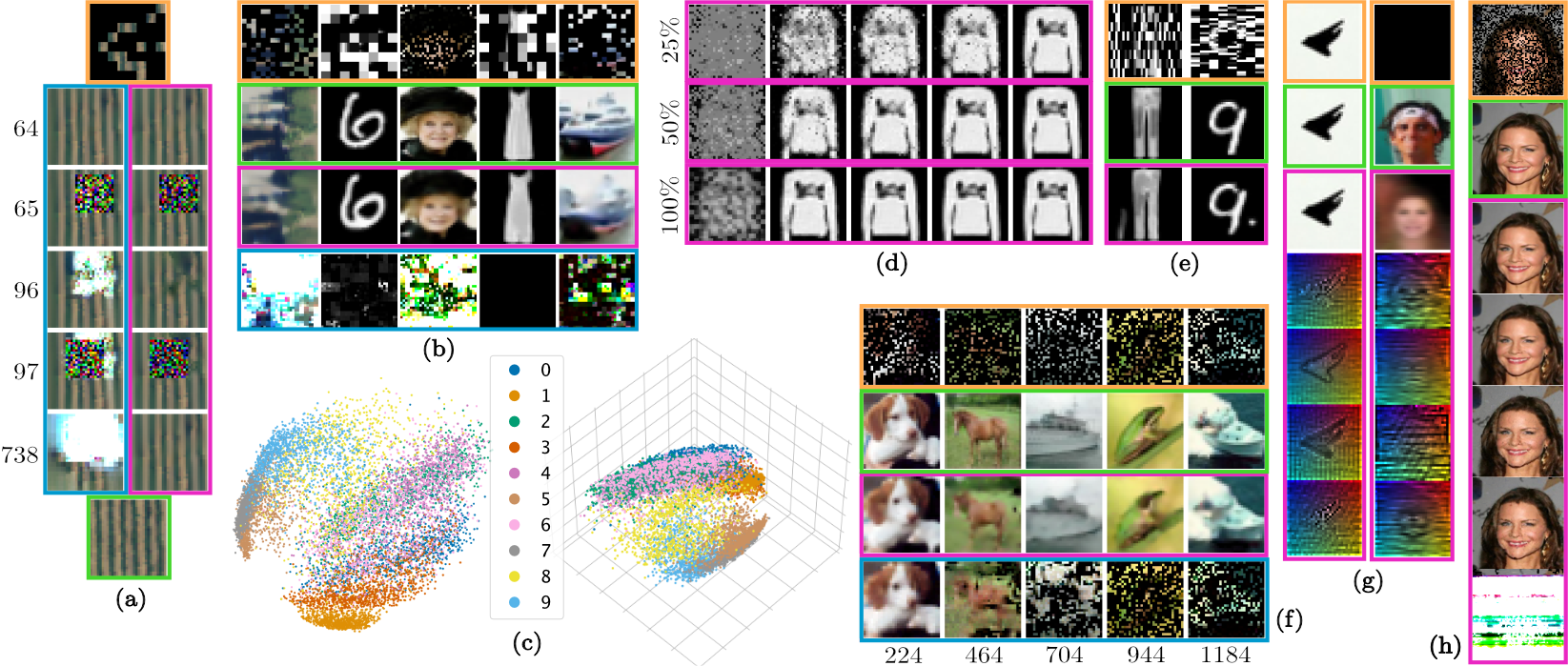}}
  \caption{Qualitative results. \textcolor{inGold}{Gold} boxes are inputs, \textcolor{gtGreen}{green} ground truths, \textcolor{vitcaPurple}{purple} ViTCA outputs, and \textcolor{unetcaBlue}{blue} UNetCA outputs. We analyze the effects of ViTCA and UNetCA on cell states in terms of: \textbf{(a)} damage resilience; \textbf{(b)} convergence stabilility, and; \textbf{(c)} hidden state PCA visualizations of converged cell grids for all examples in FashionMNIST~\cite{xiao2017fashion}. We also investigate update rule inductive biases in terms of adapting to: \textbf{(f)} varying inputs \emph{during} cell updates; \textbf{(d)} varying cell update rates; \textbf{(e)} noise configurations unseen during training; \textbf{(g)} unmasked and completely masked inputs, and; \textbf{(h)} spatial interpolation enabled by our various methods of incorporating cell positioning.\vspace{-\baselineskip}}
  \label{fig:analysis}
\end{figure}

Here we provide an empirically-based qualitative analysis on the effects ViTCA and UNetCA have on cell states through several experiments with our pre-trained models (\Fig{\ref{fig:analysis} (a,b,c)}). We notice that in general, ViTCA indefinitely maintains cell state stability while UNetCA typically induces a divergence past a certain point. An extended analysis is available in Appendix \ref{subsec:extended_analysis}.

\textbf{Damage resilience.}
Shown in \Fig{\ref{fig:analysis}} (a), we damage a random $H/2\!\times\!W/2$ patch of cells with random values $\sim\!\mathcal{U}(-1,1)$ twice in succession. ViTCA is able to maintain cell stability despite not being trained to deal with such noise, while UNetCA induces a divergence. Note both models are simultaneously performing the typical denoising task. We also note that ViTCA's inherent damage resilience is in contrast to recent NCA formulations that required explicit training for it \cite{palm2022variational,mordvintsev2020growing}.

\textbf{Convergence stability.}
\Fig{\ref{fig:analysis}} (b) shows denoising results after 2784 cell grid updates. ViTCA is able to maintain a stable cell grid state while UNetCA causes cells to diverge.

\textbf{Hidden state visualizations.}
\Fig{\ref{fig:analysis}} (c) shows 2D and 3D PCA dimensionality reductions on the hidden states of converged cell grids for all examples in FashionMNIST~\cite{xiao2017fashion}. The clusters suggest some linear separability in the learned representation, motivating our probing experiments in \Sec{\ref{subsec:probing}}. 

\subsubsection{Investigating update rule inductive biases}
\label{subsubsec:inductive_bias_investigation}
\vspace{-.5\baselineskip}

Here we investigate the inductive biases inherent in ViTCA and UNetCA by testing their adaptation to various environmental changes (\Fig{\ref{fig:analysis}} (d,e,f,g,h)).

\textbf{Adaptation to varying update rates.}
Despite being trained with a $\sigma\!=\!50\%$ cell update rate, ViTCA is able to adapt to varying rates (\Fig{\ref{fig:analysis}} (d)). Higher rates result in a proportionally faster rate of cell state convergence, and equivalently with lower rates. UNetCA exhibits a similar relationship, although is unstable at $\sigma\!=\!100\%$ (see Appendix \ref{subsec:extended_analysis}). For details comparing training with a synchronous \vs asynchronous cell grid update, see Appendix \ref{subsec:extended_ablation}.

\textbf{Generalization to noise unseen during training.}
ViTCA is capable of denoising configurations of noise it has not been trained on. \Fig{\ref{fig:analysis}} (e; \emph{left-to-right}): $4\!\times\!1$ and $1\!\times\!4$ patches of Gaussian noise at 65\% coverage. In contrast, UNetCA induces a cell state divergence (see Appendix \ref{subsec:extended_analysis}).

\textbf{Adaptation to changing inputs.}
At various moments during cell updates, we re-inject cells with new masked inputs (\Fig{\ref{fig:analysis}} (f)). ViTCA is able to consistently adapt cells to new inputs while UNetCA experiences difficulty past a certain point (\eg at 464 iterations in the figure).

\textbf{Effects of not \vs completely masking input.}
\Fig{\ref{fig:analysis}} (g; \emph{left}): ViTCA is able to perform autoencoding despite not being trained for it. UNetCA induces a cell grid divergence (see Appendix \ref{subsec:extended_analysis}). \Fig{\ref{fig:analysis}} (g; \emph{right}): Interestingly, when the input is completely masked, ViTCA outputs the median image~\cite{lehtinen2018noise2noise}. UNetCA does not exhibit such behaviour and instead causes cells to diverge (see Appendix \ref{subsec:extended_analysis}). 

\textbf{Spatial interpolation.}
We use ViTCA models trained at $32\!\times\!32$ using various types of positioning to generate $128\!\times\!128$ outputs during inference, assuming an identical cell grid resolution. \Fig{\ref{fig:analysis}} (h; \emph{top-to-bottom of \textcolor{vitcaPurple}{outputs}}): xy-coordinates, no positioning, Fourier features \cite{mildenhall2020nerf}, Fourier features concatenated with xy-coordinates, and a Transformer-based handcrafted positional encoding (baseline) \cite{vaswani2017attention}. Results are ordered from best to worst. The baseline approach is not capable of spatial interpolation due to being a 1D positioning, while, as expected, the 2D encodings make it capable. Surprisingly, removing Fourier features and using only xy-coordinates results in a higher fidelity interpolation. We believe this to be caused by the distracting amount of positional information Fourier features provide to cells, as cells can instead rely on their hidden states to store higher frequency positional information. Finally, with no explicit positioning, ViTCA is still able to perform high-quality interpolation---even exceeding using Fourier features---by taking advantage of its self-organizing nature. As a side note, we point attention to the fact that ViTCA is simultaneously denoising at a scale space it has not been trained on, exemplifying its generalization capabilities.

\vspace{-.5\baselineskip}
\subsection{Investigating hidden representations via linear probes}
\vspace{-.5\baselineskip}
\label{subsec:probing}

\begin{table}[t!]
  \caption{Linear probe \cite{chen2020generative} test accuracies (\%) of baseline and variant models. Model variants are labelled as in \Tab{\ref{tab:denoising_results}}. All baselines and variants were pre-trained for denoising autoencoding and kept fixed during probing. A linear classifier and 2-layer Multilayer Perceptrons (MLP) were trained on raw image inputs. Parameter counts exclude fixed parameters. Boldface and underlined values denote the best and second best results, respectively. Interestingly, CA-based models trained for denoising autoencoding on increasingly challenging datasets produce an increasingly more useful self-supervised representation for image classification compared to non-CA-based models. \vspace{-\baselineskip}}
  \label{tab:linear_probe_results}
  \centering
  \small
    \begin{tabular}{@{}*{8}{l}@{}}
        & & \multicolumn{2}{c}{\textbf{MNIST}} & \multicolumn{2}{c}{\textbf{FashionMNIST}} & \multicolumn{2}{c}{\textbf{CIFAR10}} \\
        \cmidrule[\heavyrulewidth](lr){3-4} \cmidrule[\heavyrulewidth](lr){5-6}
        \cmidrule[\heavyrulewidth](lr){7-8}
        \addlinespace[-\aboverulesep]
        & & \acclabel & \paramslabel & \acclabel & \paramslabel & \acclabel & \paramslabel \\
        \toprule
        \multirow{4}{*}{\rot{\textbf{Baselines}}}
        & U-Net & 96.3 & 15.4K & 86.2 & 15.4K & 52.3 & 15.4K \\
        & ViT & 92.1 & 1.3M & 83.4 & 1.3M & 34.5 & 1.3M \\
        & UNetCA & 96.3 & 327.7K & 89.5 & 327.7K & \textbf{55.1} & 327.7K \\
        & ViTCA & 96.7 & 327.7K & 89.7 & 327.7K & 50.2 & 327.7K \\
        \cdashrule{1-8}
        \multirow{6}{*}{\rot{\textbf{Variants}}}
        & ViTCA-32 & 96.3 & 327.7K & \underline{89.8} & 327.7K & \textbf{55.1} & 327.7K \\
        & ViTCA-32xy & 96.3 & 327.7K & 89.5 & 327.7K & \underline{53.6} & 327.7K \\
        & ViTCA-i & 95.8 & 327.7K & 89.6 & 327.7K & 49.4 & 327.7K \\
        & ViTCA-i16 & 95.7 & 327.7K & \textbf{90.1} & 327.7K & 50.7 & 327.7K \\
        & ViTCA-ixy & 96.2 & 327.7K & 89.6 & 327.7K & 50.2 & 327.7K  \\
        & ViTCA-i16xy & 96.5 & 327.7K & 89.6 & 327.7K & 52.7 & 327.7K \\
        \cdashrule{1-8}
        & Linear classifier & 93.0 & 10.3K & 84.7 & 10.3K & 39.0 & 30.7K \\
        & 2-layer MLP, 100 hidden units & \underline{98.2} & 103.5K & 89.4 & 103.5K & 46.0 & 308.3K \\
        & 2-layer MLP, 1000 hidden units & \textbf{98.5} & 1.0M & 89.6 & 1.0M & 49.7 & 3.1M \\
        \bottomrule
    \end{tabular}
\end{table}

Here we examine the learned representations of our models pre-trained for denoising.  We freeze model parameters and learn linear classifiers on each of their learned representations: converged cell hidden states for CA-based models, bottleneck features for U-Net, and LN'd tokens for ViT. This is a common approach used to probe learned representations~\cite{chen2020generative}. Classification results on MNIST, FashionMNIST, and CIFAR10 are shown in \Tab{\ref{tab:linear_probe_results}} and we use the same training setup as for denoising, but without any noise. For comparison, we also provide results using a linear classifier and two 2-layer MLPs of varying complexity, all trained directly on raw pixel values. Correlations between denoising performance in \Tab{\ref{tab:denoising_results}} and classification performance in \Tab{\ref{tab:linear_probe_results}} can be observed. Linear classification accuracy on ViTCA-based features typically exceeds classification accuracy using other model-based features or raw pixel values, even outperforming the MLPs in most cases.

\vspace{-.5\baselineskip}
\section{Discussion}
\label{sec:discussion}
\vspace{-.5\baselineskip}
We have performed extensive quantitative and qualitative evaluations of our newly proposed ViTCA on a variety of datasets under a denoising autoencoding framework. We have demonstrated the superior denoising performance and robustness of our model when compared to a U-Net-based CA baseline (UNetCA) and ViT, as well as its generalization capabilities under a variety of environmental changes such as larger inputs (\ie spatial interpolation) and changing inputs \emph{during} cell updates.

Despite the computation savings---owed to our circumvention of self-attention's quadratic complexity by spatially localizing it within ViTCA---there remains the same memory limitations inherent to all recurrent models: multiple recurrent iterations are required for each training iteration, resulting in larger memory usage than a feedforward approach. This limits single-GPU training accessibility. We have experimented with gradient checkpointing \cite{chen2016training} but found its trade-off for increased backpropagation duration (and slightly different gradients) less than ideal. To fully realize the potential of NCAs (self-organization, inherent distributivity, \etc), we encourage follow-up work to address this limitation. Adapting recent techniques using implicit differentiation is one avenue to circumvent these issues~\cite{bai2022deep,bai2019deep}. Also, as mentioned in our ablation (\Sec{\ref{subsubsec:ablation_study}}), we hope to further investigate the instabilities caused by increasing the depth of ViTCA.
%

\newpage
\begin{ack}
First and foremost, M.\@T.\ thanks their former supervisor and mentor, Konstantinos (Kosta) G.\ Derpanis, for his invaluable support throughout the project.
M.\@T.\ also thanks Martin Weiss for his helpful feedback on implementing the linear probe experiments (\Sec{\ref{subsec:probing}});
Olexa Bilaniuk for his assistance in investigating the gradient differences caused by PyTorch's gradient checkpointing implementation (see Appendix \ref{subsec:extended_ablation}), and;
the Mila Innovation, Development, and Technology (IDT) team for their overall technical support, particularly, their tireless efforts maintaining cluster reliability during the crucial moments preceding the submission deadline.

M.\@T.\ is partially supported by the Natural Sciences and Engineering Research Council of Canada (NSERC) Canada Graduate Scholarship -- Doctoral [application number CGSD3-519428-2018].
D.\@N.\ and C.\@P.\ are each partially supported by an NSERC Discovery Grant [application IDs 5011360 and 5018358, respectively]. D.\@N.\ thanks Samsung Electronics Co.\ Ldt.\ for their support. C.\@P.\ thanks CIFAR for their support under the AI Chairs Program.
\end{ack}

{\small
\printbibliography

@String(PAMI  = {IEEE Transactions on Pattern Analysis and Machine Intelligence (TPAMI)})

@String(IJCV  = {International Journal of Computer Vision (IJCV)})

@String(CVPR  = {IEEE/CVF Computer Vision and Pattern Recognition Conference (CVPR)})

@String(ICCV  = {IEEE/CVF International Conference on Computer Vision (ICCV)})

@String(ECCV  = {European Conference on Computer Vision (ECCV)})

@String(NeurIPS  = {Neural Information Processing Systems (NeurIPS)})

@String(TIP   = {IEEE Transactions on Image Processing (TIP)})

@String(VISUAL = {IEEE Conference on Visualization})

@String(ICLR  = {International Conference on Learning Representations (ICLR)})

@String(ICLRW  = {International Conference on Learning Representations Workshops (ICLR Workshops)})

@String(IMI = {Innovations in Machine Intelligence (IMI)})

@String(ICML = {International Conference on Machine Learning (ICML)})

@String(NAACL = {Proceedings of the North {A}merican Chapter of the Association for Computational Linguistics (NAACL): Human Language Technologies})

@String(MICCAI= {International Conference on Medical Image Computing and Computer-Assisted Intervention (MICCAI)})

@String(JMLR = {Journal of Machine Learning Research (JMLR)})

@String(JCSS = {Journal of Computer and System Sciences (JCSS)})

@String(PRE = {Physical Review E (PRE)})

@inproceedings{lehtinen2018noise2noise,
  title={Noise2Noise: Learning Image Restoration without Clean Data},
  author={Lehtinen, Jaakko and Munkberg, Jacob and Hasselgren, Jon and Laine, Samuli and Karras, Tero and Aittala, Miika and Aila, Timo},
  booktitle=ICML,
  pages={2965--2974},
  year={2018}
}

@article{mordvintsev2021mu,
  title={{μNCA}: Texture Generation with Ultra-Compact Neural Cellular Automata},
  author={Mordvintsev, Alexander and Niklasson, Eyvind},
  journal={arXiv preprint arXiv:2111.13545},
  year={2021}
}

@inproceedings{chen2020generative,
  title={Generative Pretraining From Pixels},
  author={Chen, Mark and Radford, Alec and Child, Rewon and Wu, Jeffrey and Jun, Heewoo and Luan, David and Sutskever, Ilya},
  booktitle=ICLR,
  pages={1691--1703},
  year={2020}
}

@inproceedings{zhang2020learning,
  title={Learning to Generate {3D} Shapes with Generative Cellular Automata},
  author={Zhang, Dongsu and Choi, Changwoon and Kim, Jeonghwan and Kim, Young Min},
  booktitle=ICLR,
  year={2021}
}

@article{cook2004universality,
  title={Universality in elementary cellular automata},
  author={Cook, Matthew},
  journal={Complex systems},
  pages={1--40},
  year={2004}
}

@article{siegelmann1995computational,
  title={On the computational power of neural nets},
  author={Siegelmann, Hava T and Sontag, Eduardo D},
  journal=JCSS,
  pages={132--150},
  year={1995}
}

@article{lecun1998gradient,
  title={Gradient-based learning applied to document recognition},
  author={LeCun, Yann and Bottou, L{\'e}on and Bengio, Yoshua and Haffner, Patrick},
  journal={Proceedings of the IEEE},
  pages={2278--2324},
  year={1998}
}

@inproceedings{ronneberger2015u,
  title={{U-Net}: Convolutional networks for biomedical image segmentation},
  author={Ronneberger, Olaf and Fischer, Philipp and Brox, Thomas},
  booktitle=MICCAI,
  pages={234--241},
  year={2015},
  organization={Springer}
}

@article{sandler2020image,
  title={Image segmentation via cellular automata},
  author={Sandler, Mark and Zhmoginov, Andrey and Luo, Liangcheng and Mordvintsev, Alexander and Randazzo, Ettore and others},
  journal={arXiv preprint arXiv:2008.04965},
  year={2020}
}

@inproceedings{he2016deep,
  title={Deep Residual Learning for Image Recognition},
  author={He, Kaiming and Zhang, Xiangyu and Ren, Shaoqing and Sun, Jian},
  booktitle=CVPR,
  pages={770--778},
  year={2016},
  organization={IEEE}
}

@article{khan2021transformers,
  title={Transformers in vision: A survey},
  author={Khan, Salman and Naseer, Muzammal and Hayat, Munawar and Zamir, Syed Waqas and Khan, Fahad Shahbaz and Shah, Mubarak},
  journal={ACM Computing Surveys (CSUR)},
  year={2021},
  publisher={ACM New York, NY}
}

@inproceedings{devlin-etal-2019-bert,
    title="{BERT}: Pre-training of Deep Bidirectional Transformers for Language Understanding",
    author="Devlin, Jacob  and
      Chang, Ming-Wei  and
      Lee, Kenton  and
      Toutanova, Kristina",
    booktitle=NAACL,
    year="2019",
    pages="4171--4186",
}

@article{he2021masked,
  title={Masked autoencoders are scalable vision learners},
  author={He, Kaiming and Chen, Xinlei and Xie, Saining and Li, Yanghao and Doll{\'a}r, Piotr and Girshick, Ross},
  journal={arXiv preprint arXiv:2111.06377},
  year={2021}
}

@misc{krizhevsky2009learning,
  title={Learning multiple layers of features from tiny images},
  author={Krizhevsky, Alex and Hinton, Geoffrey and others},
  year={2009},
  note={\url{https://www.cs.toronto.edu/~kriz/learning-features-2009-TR.pdf}}
}

@inproceedings{bengio2009curriculum,
  title={Curriculum learning},
  author={Bengio, Yoshua and Louradour, J{\'e}r{\^o}me and Collobert, Ronan and Weston, Jason},
  booktitle=ICML,
  pages={41--48},
  year={2009}
}

@article{wang2021survey,
  title={A survey on curriculum learning},
  author={Wang, Xin and Chen, Yudong and Zhu, Wenwu},
  journal=PAMI,
  year={2021},
}

@inproceedings{mildenhall2020nerf,
  title={{NeRF}: Representing scenes as neural radiance fields for view synthesis},
  author={Mildenhall, Ben and Srinivasan, Pratul P and Tancik, Matthew and Barron, Jonathan T and Ramamoorthi, Ravi and Ng, Ren},
  booktitle=ECCV,
  pages={405--421},
  year={2020}
}

@inproceedings{bai2022deep,
    author = {Bai, Shaojie and Geng, Zhengyang and Savani, Yash and Kolter, J. Zico},
    title = {Deep Equilibrium Optical Flow Estimation},
    booktitle = CVPR,
    year = {2022}
}

@article{von1966theory,
%   title={Theory of self-reproducing automata},
%   author={Von Neumann, John and Burks, Arthur W and others},
%   journal={IEEE Transactions on Neural Networks},
%   volume={5},
%   number={1},
%   pages={3--14},
%   year={1966}
% }

@inproceedings{perez2019turing,
  title={On the turing completeness of modern neural network architectures},
  author={P{\'e}rez, Jorge and Marinkovi{\'c}, Javier and Barcel{\'o}, Pablo},
  booktitle=ICLR,
  year={2018}
}

@article{cordonnier2019relationship,
  title={On the relationship between self-attention and convolutional layers},
  author={Cordonnier, Jean-Baptiste and Loukas, Andreas and Jaggi, Martin},
  booktitle=ICLR,
  year={2019}
}

@article{ba2016layer,
  title={Layer normalization},
  author={Ba, Jimmy Lei and Kiros, Jamie Ryan and Hinton, Geoffrey E},
  journal={arXiv preprint arXiv:1607.06450},
  year={2016}
}

@article{chen2016training,
  title={Training deep nets with sublinear memory cost},
  author={Chen, Tianqi and Xu, Bing and Zhang, Chiyuan and Guestrin, Carlos},
  journal={arXiv preprint arXiv:1604.06174},
  year={2016}
}

@article{loshchilov2017sgdr,
  title={{SGDR}: Stochastic gradient descent with warm restarts},
  author={Loshchilov, Ilya and Hutter, Frank},
  journal=ICLR,
  year={2017}
}

@inproceedings{loshchilov2019decoupled,
  title={Decoupled weight decay regularization},
  author={Loshchilov, Ilya and Hutter, Frank},
  booktitle=ICLR,
  year={2019}
}

@article{russakovsky2015imagenet,
  author={Olga Russakovsky and Jia Deng and Hao Su and Jonathan Krause and Sanjeev Satheesh and Sean Ma and Zhiheng Huang and Andrej Karpathy and Aditya Khosla and Michael Bernstein and Alexander C. Berg and Li Fei-Fei},
  title={{ImageNet Large Scale Visual Recognition Challenge}},
  year={2015},
  journal=IJCV,
  pages={211-252}
}

@article{wei2021masked,
  title={Masked Feature Prediction for Self-Supervised Visual Pre-Training},
  author={Wei, Chen and Fan, Haoqi and Xie, Saining and Wu, Chao-Yuan and Yuille, Alan and Feichtenhofer, Christoph},
  journal={arXiv preprint arXiv:2112.09133},
  year={2021}
}

@article{niklasson2021self-organising,
  author = {Niklasson, Eyvind and Mordvintsev, Alexander and Randazzo, Ettore and Levin, Michael},
  title = {Self-Organising Textures},
  journal = {Distill},
  year = {2021},
  note = {\url{https://distill.pub/selforg/2021/textures}}
}

@article{vincent2010stacked,
  title={Stacked denoising autoencoders: Learning useful representations in a deep network with a local denoising criterion.},
  author={Vincent, Pascal and Larochelle, Hugo and Lajoie, Isabelle and Bengio, Yoshua and Manzagol, Pierre-Antoine and Bottou, L{\'e}on},
  journal=JMLR,
  year={2010}
}

@inproceedings{jaegle2021perceiver,
  title={Perceiver {IO}: A general architecture for structured inputs \& outputs},
  author={Jaegle, Andrew and Borgeaud, Sebastian and Alayrac, Jean-Baptiste and Doersch, Carl and Ionescu, Catalin and Ding, David and Koppula, Skanda and Zoran, Daniel and Brock, Andrew and Shelhamer, Evan and others},
  booktitle=ICLR,
  year={2022}
}

@inproceedings{yifan2021input,
  title={Input-level Inductive Biases for 3D Reconstruction},
  author={Yifan, Wang and Doersch, Carl and Arandjelovi{\'c}, Relja and Carreira, Jo{\~a}o and Zisserman, Andrew},
  booktitle=CVPR,
  year={2022}
}

@inproceedings{ali2021xcit,
  title={{XCiT}: Cross-covariance image transformers},
  author={Ali, Alaaeldin and Touvron, Hugo and Caron, Mathilde and Bojanowski, Piotr and Douze, Matthijs and Joulin, Armand and Laptev, Ivan and Neverova, Natalia and Synnaeve, Gabriel and Verbeek, Jakob and others},
  booktitle=NeurIPS,
  year={2021}
}

@inproceedings{variengien2021towards,
  title={Towards self-organized control: Using neural cellular automata to robustly control a cart-pole agent},
  author={Variengien, Alexandre and Nichele, Stefano and Glover, Tom and Pontes-Filho, Sidney},
  booktitle=IMI,
  pages={1--14},
  year={2021}
}

@inproceedings{hudson2021generative,
  title={Generative adversarial transformers},
  author={Hudson, Drew A and Zitnick, Larry},
  booktitle=ICML,
  pages={4487--4499},
  year={2021}
}

@inproceedings{arnab2021vivit,
  title={{ViViT}: A video vision transformer},
  author={Arnab, Anurag and Dehghani, Mostafa and Heigold, Georg and Sun, Chen and Lu{\v{c}}i{\'c}, Mario and Schmid, Cordelia},
  booktitle=ICCV,
  pages={6836--6846},
  year={2021}
}

@inproceedings{crawfis1993texture,
  title={Texture splats for {3D} scalar and vector field visualization},
  author={Crawfis, Roger A and Max, Nelson},
  booktitle=VISUAL,
  pages={261--266},
  year={1993}
}

@inproceedings{fan2021multiscale,
  title={Multiscale vision transformers},
  author={Fan, Haoqi and Xiong, Bo and Mangalam, Karttikeya and Li, Yanghao and Yan, Zhicheng and Malik, Jitendra and Feichtenhofer, Christoph},
  booktitle=ICCV,
  pages={6824--6835},
  year={2021}
}

@inproceedings{vaswani2017attention,
  title={Attention is all you need},
  author={Vaswani, Ashish and Shazeer, Noam and Parmar, Niki and Uszkoreit, Jakob and Jones, Llion and Gomez, Aidan N and Kaiser, {\L}ukasz and Polosukhin, Illia},
  booktitle=NeurIPS,
  year={2017}
}

@inproceedings{dosovitskiy2020image,
  title={An Image is Worth 16x16 Words: Transformers for Image Recognition at Scale},
  author={Dosovitskiy, Alexey and Beyer, Lucas and Kolesnikov, Alexander and Weissenborn, Dirk and Zhai, Xiaohua and Unterthiner, Thomas and Dehghani, Mostafa and Minderer, Matthias and Heigold, Georg and Gelly, Sylvain and Uszkoreit, Jakob and Houlsby, Neil},
  booktitle=ICLR,
  year={2021}
}

@inproceedings{najarro2022hypernca,
  title={Hyper{NCA}: Growing Developmental Networks with Neural Cellular Automata},
  author={Elias Najarro and Shyam Sudhakaran and Claire Glanois and Sebastian Risi},
  booktitle=ICLRW,
  year={2022}
}

@inproceedings{palm2022variational,
  title={Variational Neural Cellular Automata},
  author={Rasmus Berg Palm and Miguel Gonz{\'a}lez Duque and Shyam Sudhakaran and Sebastian Risi},
  booktitle=ICLR,
  year={2022},
  url={https://openreview.net/forum?id=7fFO4cMBx_9}
}

@inproceedings{grattarola2021learning,
  title={Learning Graph Cellular Automata},
  author={Grattarola, Daniele and Livi, Lorenzo and Alippi, Cesare},
  booktitle=NeurIPS,
  pages = {20983--20994},
  year={2021}
}

@article{mordvintsev2020growing,
  author = {Mordvintsev, Alexander and Randazzo, Ettore and Niklasson, Eyvind and Levin, Michael},
  title = {Growing Neural Cellular Automata},
  journal = {Distill},
  year = {2020},
  note = {\url{https://distill.pub/2020/growing-ca}}
}

@inproceedings{wulff1992learning,
  title={Learning cellular automaton dynamics with neural networks},
  author={Wulff, N H and Hertz, J A},
  booktitle=NeurIPS,
  pages={631--638},
  year={1992}
}

@inproceedings{yeh2021sustainbench,
  title = {{SustainBench}: Benchmarks for Monitoring the Sustainable Development Goals with Machine Learning},
  author = {Christopher Yeh and Chenlin Meng and Sherrie Wang and Anne Driscoll and Erik Rozi and Patrick Liu and Jihyeon Lee and Marshall Burke and David Lobell and Stefano Ermon},
  booktitle = NeurIPS,
  year = {2021}
}

@inproceedings{chen2022regionvit,
  title={{RegionViT}: Regional-to-Local Attention for Vision Transformers},
  author={Chun-Fu Chen and Rameswar Panda and Quanfu Fan},
  booktitle=ICML,
  year={2022},
  url={https://openreview.net/forum?id=T__V3uLix7V}
}

@inproceedings{liu2021swin,
  title={Swin transformer: Hierarchical vision transformer using shifted windows},
  author={Liu, Ze and Lin, Yutong and Cao, Yue and Hu, Han and Wei, Yixuan and Zhang, Zheng and Lin, Stephen and Guo, Baining},
  booktitle=ICCV,
  pages={10012--10022},
  year={2021}
}

@inproceedings{chu2021twins,
  title={Twins: Revisiting spatial attention design in vision transformers},
  author={Chu, Xiangxiang and Tian, Zhi and Wang, Yuqing and Zhang, Bo and Ren, Haibing and Wei, Xiaolin and Xia, Huaxia and Shen, Chunhua},
  booktitle=NeurIPS,
  year={2021}
}

@inproceedings{zhang2021multi,
  title={Multi-Scale Vision Longformer: A New Vision Transformer for High-Resolution Image Encoding},
  author= {Zhang, Pengchuan and Dai, Xiyang and Yang, Jianwei and Xiao, Bin and Yuan, Lu and Zhang, Lei and Gao, Jianfeng},
  booktitle=ICCV,
  year={2021},
  pages={2998-3008}
}

@inproceedings{liu2015faceattributes,
  title = {Deep Learning Face Attributes in the Wild},
  author = {Liu, Ziwei and Luo, Ping and Wang, Xiaogang and Tang, Xiaoou},
  booktitle = ICCV,
  pages={3730--3738},
  year = {2015} 
}

@article{bai2019deep,
  title={Deep equilibrium models},
  author={Bai, Shaojie and Kolter, J Zico and Koltun, Vladlen},
  journal=NeurIPS,
  year={2019}
}

@article{deng2012mnist, 
  title={The {MNIST} database of handwritten digit images for machine learning research}, 
  author={Deng, Li}, 
  journal={IEEE Signal Processing Magazine}, 
  pages={141--142}, 
  year={2012}
}

@article{xiao2017fashion,
  title={Fashion-{MNIST}: a novel image dataset for benchmarking machine learning algorithms},
  author={Xiao, Han and Rasul, Kashif and Vollgraf, Roland},
  journal={arXiv preprint arXiv:1708.07747},
  year={2017}
}

@inproceedings{he2015delving,
  title={Delving deep into rectifiers: Surpassing human-level performance on imagenet classification},
  author={He, Kaiming and Zhang, Xiangyu and Ren, Shaoqing and Sun, Jian},
  booktitle=ICCV,
  pages={1026--1034},
  year={2015}
}

@inproceedings{zhang2018unreasonable,
  title={The unreasonable effectiveness of deep features as a perceptual metric},
  author={Zhang, Richard and Isola, Phillip and Efros, Alexei A and Shechtman, Eli and Wang, Oliver},
  booktitle=CVPR,
  pages={586--595},
  year={2018}
}

@inproceedings{wang2004image,
  title={Image quality assessment: from error visibility to structural similarity},
  author={Wang, Zhou and Bovik, Alan C and Sheikh, Hamid R and Simoncelli, Eero P},
  booktitle=TIP,
  pages={600--612},
  year={2004}
}

@misc{weissteinmoore,
  author={Weisstein, Eric W.},
  title={Moore Neighborhood. {From MathWorld---A Wolfram Web Resource}},
  url={https://mathworld.wolfram.com/MooreNeighborhood.html}
}

@article{gilpin2019cellular,
  title={Cellular automata as convolutional neural networks},
  author={Gilpin, William},
  journal=PRE,
  volume={100},
  issue={3},
  pages={032402},
  numpages={11},
  year={2019},
  publisher={American Physical Society (APS)}
}

@inproceedings{bao2022beit,
  title={{BEiT}: {BERT} Pre-Training of Image Transformers},
  author={Hangbo Bao and Li Dong and Songhao Piao and Furu Wei},
  booktitle=ICLR,
  year={2022},
  url={https://openreview.net/forum?id=p-BhZSz59o4}
}

@incollection{christen2021automatic,
  title={Automatic programming of cellular automata and artificial neural networks guided by philosophy},
  author={Christen, Patrik and Del Fabbro, Olivier},
  booktitle={New Trends in Business Information Systems and Technology},
  pages={131--146},
  year={2021}
}
}

\section*{Checklist}
\begin{enumerate}

\item For all authors...
\begin{enumerate}
  \item Do the main claims made in the abstract and introduction accurately reflect the paper's contributions and scope?
    \answerYes
  \item Did you describe the limitations of your work?
    \answerYes{See \Sec{\ref{sec:discussion}}.}
  \item Did you discuss any potential negative societal impacts of your work?
    \answerNo{Although we feel our work demonstrates the potential of NCAs as viable alternatives to common recurrent network architectures (ViTCA being our evidential contribution), our experiments intentionally tend towards the direction of optimizing model efficiency (and single-GPU training accessibility) rather than towards the increasingly popular direction of scaling upwards. However, as much as our work demonstrates the downward-scaling capabilities of NCAs, we also acknowledge that this similarly applies going upward, and as such, can be abused (\eg creating a ``deepfake''-capable ViTCA).}
  \item Have you read the ethics review guidelines and ensured that your paper conforms to them?
    \answerYes
\end{enumerate}

\item If you are including theoretical results...
\begin{enumerate}
  \item Did you state the full set of assumptions of all theoretical results
    \answerNA{}
        \item Did you include complete proofs of all theoretical results?
    \answerNA{}
\end{enumerate}

\item If you ran experiments...
\begin{enumerate}
  \item Did you include the code, data, and instructions needed to reproduce the main experimental results (either in the supplemental material or as a URL)?
    \answerYes{Code and instructions to reproduce results are included in the supplemental material.}
  \item Did you specify all the training details (e.g., data splits, hyperparameters, how they were chosen)?
    \answerYes{}
        \item Did you report error bars (e.g., with respect to the random seed after running experiments multiple times)?
    \answerNo{Given the combination of time and computational restrictions and our exhaustive list of experiments, we opted to prioritize experiment variety and dataset coverage as an implicit substitute for re-running experiments under different random seeds. For all experiments, we kept a fixed random seed, even pointing out (deterministic) differences caused by gradient checkpointing when used (see Appendix \ref{sec:appendix}).}
        \item Did you include the total amount of compute and the type of resources used (e.g., type of GPUs, internal cluster, or cloud provider)?
    \answerYes
\end{enumerate}

\item If you are using existing assets (e.g., code, data, models) or curating/releasing new assets...
\begin{enumerate}
  \item If your work uses existing assets, did you cite the creators?
    \answerYes
  \item Did you mention the license of the assets?
    \answerNA{Licensed frameworks used such as PyTorch (BSD-style) and Hydra (MIT) will be mentioned in acknowledgements.}
  \item Did you include any new assets either in the supplemental material or as a URL?
    \answerNo{No new assets---aside from code and training our models---were created for the purposes of this work.}
  \item Did you discuss whether and how consent was obtained from people whose data you're using/curating?
    \answerNo{We used publicly available datasets.}
  \item Did you discuss whether the data you are using/curating contains personally identifiable information or offensive content?
    \answerNo{Although not discussed in the manuscript, we would like to point that the datasets we used that could potentially contain personally identifiable information (CelebA, CIFAR10, Tiny ImageNet) each have restrictions and/or acknowledgements of such potential issues. Also, our work is not focused on classifying persons and ViTCA is not a generative model, \eg it can not generate new faces.}
\end{enumerate}

\item If you used crowdsourcing or conducted research with human subjects...
\begin{enumerate}
  \item Did you include the full text of instructions given to participants and screenshots, if applicable?
    \answerNA
  \item Did you describe any potential participant risks, with links to Institutional Review Board (IRB) approvals, if applicable?
    \answerNA
  \item Did you include the estimated hourly wage paid to participants and the total amount spent on participant compensation?
    \answerNA
\end{enumerate}

\end{enumerate}

\newpage
\appendix
\section{Appendix}
\label{sec:appendix}

\begin{algorithm}[H]
    \label{alg:vitca_training}
    \caption{Training the ViTCA update rule with a ``pool sampling''-based approach}
    \Input{ViTCA cell update rule $F_\theta$, hyper-parameters $\Omega\!=\!$ $\{I\in\mathbb{N}^+,$ $b\in\mathbb{N}^+,$ $\sigma \in [0,1],$ $C_h \in \mathbb{N}^+,$ $\eta \in \mathbb{R}^+,$ $\alpha \in \mathbb{R},$ $\beta \in \mathbb{R},$ $N_\mathcal{P} \in\mathbb{N}^+\}$, dataset of images $\mathcal{D}\!=\!\{\mathbf{X}_1,\mathbf{X}_2,...,$ $\mathbf{X}_{N_b\in\mathbb{N}^+}\}$}
    \Output{Optimal update rule parameters $\theta_{I}$}
    $\theta_{0} \gets$ initial update rule parameters\Comment*[r]{\Eg He initialization \cite{he2015delving}}
    $\mathcal{P} \gets \emptyset$\Comment*[r]{Pool of cell grids and their respective ground truth images}
    \For{$i\gets1$ \KwTo $I$}{
        $\mathbf{X} \gets (\mathbf{X}_j,...,\mathbf{X}_{j+b})$ where $j \sim \mathcal{U}\{1,N_b-b\}$\Comment*[r]{$(\cdot,...,\cdot)$ is batch-wise concatenation}
        \eIf{$|\mathcal{P}| > b$ \and $i \bmod 2 = 0$}{
            $\mathbf{P} \gets \{(\mathbf{Z}_1,\mathbf{X}_1),..., (\mathbf{Z}_b,\mathbf{X}_b)\} \subset \mathcal{P}$\Comment*[r]{Retrieve first $b$ elements from the pool}
            $\mathbf{Z} \gets (\mathbf{Z}_1,...,\mathbf{Z}_b);\,\,\mathbf{X} \gets (\mathbf{X}_1,...,\mathbf{X}_b)$\Comment*[r]{Retrieve cell grids and images from $\mathbf{P}$}
        }{
            \Comment{Zero-initialize grids of cells and inject noisy inputs}
            \Comment{i is used for determining noise shape and coverage}
            \Comment{$C_h$ determines the number of cell hidden channels}
            $\mathbf{Z} \gets \texttt{seed}(\texttt{mask}(\mathbf{X}, i), C_h)$\;
        }
        $T \sim \mathcal{U}\{8,32\}$\Comment*[r]{Randomly sample number of cell updates to perform}
        \For{$t\gets1$ \KwTo $T$}{
            $\mathbf{Z} \gets F_{\theta_{i-1}}(\mathbf{Z}, \sigma)$\Comment*[r]{Iteratively update cell grids with cell update prob.\ $\sigma$}
        }
        \Comment{$\mathbf{Z}_o$ and $\mathbf{Z}_h$ are output and hidden channels of cell grids, respectively}
        $L_{o\_overflow} \gets \frac{1}{C_o}\|\mathbf{Z}_o-\min(\max(\mathbf{Z}_o, 0), 1)\|_1$\Comment*[r]{Output channels overflow loss}
        $L_{h\_overflow} \gets \frac{1}{C_h}\|\mathbf{Z}_h-\min(\max(\mathbf{Z}_h, -1), 1)\|_1$\Comment*[r]{Hidden channels overflow loss}
        $L_{rec} \gets \frac{1}{C_o}\|\mathbf{Z}_o - \mathbf{X}\|_1$\Comment*[r]{Image reconstruction loss}
        $L \gets \frac{1}{bHW}(\alpha L_{rec} + \beta(L_{o\_overflow} + L_{h\_overflow}))$\;
        $Q \gets \nabla L / (\|\nabla L\|_F + 10^{-8})$\Comment*[r]{Normalize gradients.\ $\|\cdot\|_F$ is Frob.\ norm}
        $\theta_i \gets \theta_{i-1} - \eta Q$\Comment*[r]{Update the update rule parameters}
        $\mathcal{P} \gets \mathcal{P} \cup \{(\mathbf{Z}_1,\mathbf{X}_1),...,(\mathbf{Z}_b,\mathbf{X}_b)\}$\Comment*[r]{Append updated cell grids and ground truths}
        $\mathcal{P} \gets \texttt{trunc}(\texttt{shuffle}(\mathcal{P}), N_\mathcal{P})$\Comment*[r]{Shuffle pool and retain first $N_\mathcal{P}$ elements}
    }
\end{algorithm}

\subsection{Training on high-resolution imagery with fusion and mitosis}
\label{subsec:high_resolution_training}

As an alternative to gradient checkpointing for reducing memory usage, we briefly experimented with a downsampling scheme inspired by cell fusion and mitosis when training on CelebA at $64\!\times\!64$. Specifically, we split the $T$ applications of the update rule (within a training iteration) into multiple stages: 1) We apply the update rule twice so that cells will have, at minimum, some amount of knowledge of their neighbours. 2) We stash the masked input for a later re-injection. 3) \emph{Fusion}---we apply a $2\!\times\!2$ average pooling with a stride of 2 across the cell grid, combining $2\!\times\!2$ groups of cells into singular cells. 4) We apply the update rule $T-4$ times at this $32\!\times\!32$ downsampled cell grid resolution. 5) \emph{Mitosis}---we perform a $2\!\times\!2$ duplication of cells (each cell is duplicated to its right, bottom-right, and bottom). 6) We re-inject the stashed masked input. 7) We apply the update rule twice to adapt the cells to the $64\!\times\!64$ resolution and to fill in any missing information.

We found that performing this fusion and mitosis scheme decreased training memory consumption to levels similar to our gradient checkpointing scheme ($\sim\!50\%$ memory reduction) while having a $\sim\!70\%$ faster backward pass. Loss-wise, we observed a $\sim\!33\%$ increase in the average validation reconstruction loss during training, which can qualitatively be observed in the example provided in \Fig{\ref{fig:fusion_mitosis}} (\emph{\textcolor{unetcaBlue}{bottom}}). Although the results shown are not ideal---\ie we did not perform a hyper-parameter search here, for example, finding the optimal number of iterations preceding fusion and following mitosis---this brief experiment tests the feasibility of reducing memory consumption while maintaining denoising capability and avoiding gradient checkpointing. As shown in the figure, ViTCA with fusion and mitosis is able to successfully denoise the input despite applying updates at two different scales. This scale agnostic behaviour reveals potentially interesting research directions beyond the scope of this work, such as allowing an NCA update rule to dynamically and locally modify cell grid resolution based on a compute budget, which could see applications in signal (image, video, or audio) compression.

\begin{figure}[t]
  \begin{floatrow}
    \input{figures/fusion_mitosis}
    \capbtabbox[0.56\textwidth]{
  \centering
  \small
  \setlength{\tabcolsep}{3pt}
    \begin{tabular}{@{}*{6}{l}@{}}
        & & \psnrlabel & \ssimlabel & \lpipslabel & \paramslabel \\
        \toprule
        \multirow{7}{*}{\textbf{Pool size}}
        & 128 & 26.51 & \underline{0.914} & \underline{0.065} & 92.5K \\
        & 256 & 26.40 & 0.912 & 0.067 & 92.5K \\
        & 512 & \textbf{26.61} & \textbf{0.915} & \textbf{0.064} & 92.5K \\
        & \textit{1024} & 26.53 & 0.913 & 0.066 & 92.5K \\
        & 2048 & \underline{26.54} & \textbf{0.915} & \textbf{0.064} & 92.5K\\
        & 4096 & 26.48 & 0.912 & 0.066 & 92.5K \\
        & 8192 & 26.30 & 0.910 & 0.069 & 92.5K \\
        \midrule[\heavyrulewidth]
        \multirow{2}{*}{\textbf{Cell init.}}
        & \textit{constant} & \textbf{26.53} & \textbf{0.913} & \textbf{0.066} & 92.5K \\
        & random & \underline{25.90} & \underline{0.905} & \underline{0.074} & 92.5K \\
        \midrule[\heavyrulewidth]
        \multirow{5}{*}{\textbf{Patch size}}
        & \textit{1$\times$1} & \textbf{26.53} & \textbf{0.913} & \textbf{0.066} & 92.5K \\
        & 2$\times$2 & \underline{25.85} & \underline{0.906} & \underline{0.076} & 96.0K \\
        & 4$\times$4 & 24.54 & 0.882 & 0.113 & 109.8K \\
        & 8$\times$8 & 21.62 & 0.803 & 0.212 & 165.3K \\
        & 16$\times$16 & 18.71 & 0.687 & 0.279 & 387.0K \\
        \bottomrule
    \end{tabular}
}{
  \caption{Quantitative ablation on pool size $N_\mathcal{P}$, cell initialization method, and patch size $P_H\times P_W$ for denoising autoencoding with ViTCA on CelebA. Boldface and underlining denote best and second best results. Italicized items denote baseline configuration settings.}
  \label{tab:other_ablation_results}
}
    \vspace{-0.5\baselineskip}
  \end{floatrow}
\end{figure}

\subsection{Extended ablation study}
\label{subsec:extended_ablation}

Here we present an extension of our ablation study in \Sec{\ref{subsubsec:ablation_study}}, using the baseline ViTCA model as our reference. As before, the ablation examines the effects certain training configuration parameters have on test performance.

\begin{table}[t!]
  \caption{Quantitative ablation on attention neighbourhood size $N_H\times N_W$ for denoising autoencoding with ViTCA on FashionMNIST. Boldface and underlining denote best and second best results. Italicized items denote baseline configuration settings.}
  \label{tab:attn_size_ablation_results}
  \centering
  \small
  \setlength{\tabcolsep}{3pt}
    \begin{tabular}{@{}*{4}{l}@{}}
        & \psnrlabel & \ssimlabel & \lpipslabel \\
        \toprule
        \textit{3$\times$3} & \textbf{23.25} & \textbf{0.827} & \textbf{0.145} \\
        5$\times$5 & \underline{22.34} & \underline{0.817} & \textbf{0.145} \\
        7$\times$7 & 21.65 & 0.792 & \underline{0.168} \\
        \bottomrule
    \end{tabular}
\end{table}
\begin{table}[t!]
  \caption{Quantitative ablation comparing test results with ViTCA trained using asynchronous ($\sigma\!=\!50\%$) \vs\ synchronous ($\sigma\!=\!100\%$) cell updates for denoising autoencoding. During testing, cells are updated at the rate they were trained in. Boldface denotes best results. Italicized items denote baseline configuration settings.\vspace{-0.5\baselineskip}}
  \label{tab:update_rate_ablation_results}
  \centering
  \small
  \setlength{\tabcolsep}{3pt}
    \begin{tabular}{@{}*{13}{l}@{}}
        & \multicolumn{3}{c}{\textbf{LandCoverRep}} & \multicolumn{3}{c}{\textbf{MNIST}} & \multicolumn{3}{c}{\textbf{CelebA}} & \multicolumn{3}{c}{\textbf{FashionMNIST}} \\
        \cmidrule[\heavyrulewidth](lr){2-4} \cmidrule[\heavyrulewidth](lr){5-7} \cmidrule[\heavyrulewidth](lr){8-10} \cmidrule[\heavyrulewidth](lr){11-13}
        \addlinespace[-\aboverulesep]
        & \diag{\psnrlabel} & \diag{\ssimlabel} & \diag{\lpipslabel} & \diag{\psnrlabel} & \diag{\ssimlabel} & \diag{\lpipslabel} & \diag{\psnrlabel} & \diag{\ssimlabel} & \diag{\lpipslabel} & \diag{\psnrlabel} & \diag{\ssimlabel} & \diag{\lpipslabel} \\
        \toprule
        \textit{asynchronous} & \textbf{33.80} & \textbf{0.932} & \textbf{0.102} & \textbf{27.01} & \textbf{0.940} & \textbf{0.028} & \textbf{26.53} & \textbf{0.913} & \textbf{0.066} & \textbf{23.80} & \textbf{0.855} & \textbf{0.117} \\
        synchronous & 33.68 & 0.931 & 0.104 & 26.00 & 0.927 & 0.034 & 23.76 & 0.870 & 0.105 & 23.12 & 0.832 & 0.132 \\
        \bottomrule
    \end{tabular}
\end{table}
\begin{table}[t!]
  \caption{Quantitative ablation comparing test results of ViTCA trained with gradient checkpointing disabled \vs\ enabled. Boldface denotes best results. Italicized items denote baseline configuration settings.\vspace{-0.5\baselineskip}}
  \label{tab:gradient_checkpointing_ablation_results}
  \centering
  \small
  \setlength{\tabcolsep}{3pt}
    \begin{tabular}{@{}*{13}{l}@{}}
        & \multicolumn{3}{c}{\textbf{LandCoverRep}} & \multicolumn{3}{c}{\textbf{MNIST}} & \multicolumn{3}{c}{\textbf{CelebA}} & \multicolumn{3}{c}{\textbf{FashionMNIST}} \\
        \cmidrule[\heavyrulewidth](lr){2-4} \cmidrule[\heavyrulewidth](lr){5-7} \cmidrule[\heavyrulewidth](lr){8-10} \cmidrule[\heavyrulewidth](lr){11-13}
        \addlinespace[-\aboverulesep]
        & \diag{\psnrlabel} & \diag{\ssimlabel} & \diag{\lpipslabel} & \diag{\psnrlabel} & \diag{\ssimlabel} & \diag{\lpipslabel} & \diag{\psnrlabel} & \diag{\ssimlabel} & \diag{\lpipslabel} & \diag{\psnrlabel} & \diag{\ssimlabel} & \diag{\lpipslabel} \\
        \toprule
        \textit{disabled} & \textbf{33.80} & 0.932 & 0.102 & \textbf{27.01} & \textbf{0.940} & \textbf{0.028} & \textbf{26.53} & \textbf{0.913} & \textbf{0.066} & 23.80 & 0.855 & 0.117 \\
        enabled & 33.76 & 0.932 & 0.102 & 26.77 & 0.938 & 0.029 & 26.05 & 0.904 & 0.075 & \textbf{23.89} & 0.855 & 0.117 \\
        \bottomrule
    \end{tabular}
\end{table}

\paragraph{Pool size, cell initialization, and patch size.}
In \Tab{\ref{tab:other_ablation_results}}, we examine the impact of varying the (max) pool size $N_\mathcal{P}$, cell initialization method, and patch size $P_H\!\times\!P_W$ on CelebA. As shown in the table, it is difficult to correlate pool size with test performance. However, when pool size $N_\mathcal{P}\!=\!8192$, there is a noticeable reduction in performance. Test performance also degrades when initializing cells such that their output and hidden channels receive random values sampled from $\mathcal{U}(0,1)$ and $\mathcal{U}(-1,1)$, respectively, as opposed to receiving constant values (0.5 for output channels and 0 for hidden). Finally, we see a consistent decrease in performance when the input image is divided into non-overlapping patches $> 1\!\times\!1$, as well as an increase in the number of model parameters.

\paragraph{Attention neighbourhood size.}
In \Tab{\ref{tab:attn_size_ablation_results}}, we examine the impact of attention neighbourhood size $N_H\!\times\!N_W$ on FashionMNIST. Interestingly, increasing the neighbourhood size past $3\!\times\!3$ causes a degradation in performance. This is most likely attributed to the increase in complexity caused by incorporating more information into ViTCA's self-attention. One would expect explicitly increasing the receptive field of spatially localized self-attention to result in better performance, but it can also complicate the process of figuring out which neighbours to attend to. We believe this may be alleviated by increasing model capacity and/or training duration. As described in \Sec{\ref{sec:experiments}}, we use the Moore neighbourhood ($3\!\times\!3$) as it requires less computation while still demonstrating ViTCA's effectiveness.

\paragraph{Asynchronous \vs synchronous cell updates.}
In \Tab{\ref{tab:update_rate_ablation_results}}, we compare between training with asynchronous cell updates ($\sigma\!=\!50\%$) and training with synchronous cell updates ($\sigma\!=\!100\%$) on LandCoverRep, MNIST, CelebA, and FashionMNIST. Training with asynchronous cell updates provides a meaningful increase in performance compared to training with synchronous cell updates and comes with several benefits, such as not requiring cells in a neighbourhood to be in sync with each other and serving as additional data augmentation. Similarly mentioned in related work \cite{niklasson2021self-organising}, this allows ViTCA to be used in a distributed system where cells need not exist under a global clock and can be updated at varying rates. Thus making it easier to scale up or down within a non-homogeneous compute environment. This was somewhat demonstrated in \Fig{\ref{fig:analysis}} (d) where ViTCA was able to adapt to varying update rates despite being trained on a fixed asynchronous update rate ($\sigma\!=\!50\%$).

\paragraph{Effects of gradient checkpointing.}
In \Tab{\ref{tab:gradient_checkpointing_ablation_results}}, we compare between training with gradient checkpointing disabled and with gradient checkpointing enabled on LandCoverRep, MNIST, CelebA, and FashionMNIST. Similarly shown in \Tab{\ref{tab:ablation_results}}, we see here that training with gradient checkpointing has an adverse effect on test performance. As mentioned in \Sec{\ref{sec:discussion}}, NCAs---during training---require all activations from each recurrent iteration to be stored in memory before performing backpropagation. This results in memory usage being proportional to the amount of recurrent iterations. As such, depending on ViTCA's configuration, gradient checkpointing may be required to be able to train on a single GPU. We make use of PyTorch's \texttt{checkpoint\_sequential}, which we use as follows: given the number of CA iterations $T$, we divide the sequential (forward) application of the update rule into $\lfloor T/2 \rfloor$ segments of roughly the same length (depending on whether $T$ is even or odd). Then, all segments are executed in sequence, where activations from only the first and last segments are stored as well as the inputs to each intermediate segment. The intermediate inputs are used for re-running the segments without stored activations during the backward pass to compute gradients. This results in a trade-off between memory consumption and backpropagation duration since each intermediate segment's forward pass needs to be re-computed during its backward pass. Moreover, and not mentioned in the documentation of PyTorch at the time of writing, there exists a subtle yet meaningful side-effect which we have observed and confirmed through the use of GNU Debugger (GDB) and Python Debugger (PDB): Without gradient checkpointing, gradients are accumulated all at once at the end of backpropagating through the entire computation graph, resulting in the expected round-offs due to limitations in machine precision (\texttt{float32} in our case). At this point, PyTorch may use a variety of numerical techniques to minimize round-off, such as cascade summation (verified to be used for CPU-based summation, see \texttt{SumKernel.cpp} in PyTorch) which recursively sums two halves of a sequence of summands as opposed to naively summing them in sequence. \emph{With} gradient checkpointing, gradients are accumulated at each segment. This means that round-offs are forced to (potentially) occur at each checkpoint/segment instead of once at the end of the entire computation graph. Even if cascade summation is used when summing gradients within each segment, the segment-wise ordering may reduce its effectiveness. We verified this behaviour by observing an exact machine epsilon difference ($\epsilon\approx1.19\!\times\!10^{-7}$ in IEEE 754 standard) in the gradient---when compared to the non-checkpointed scheme---of the final operation of the update rule at the second-last segment, once the loss started to diverge.

It is important to note that despite the difference in gradients, the accuracy of the forward pass remains unchanged between the checkpointed and non-checkpointed models. Also, we must remind ourselves that round-offs are unavoidable when performing floating-point arithmetic, meaning that gradients computed within a deep learning library such as PyTorch are always an \emph{estimation} of the true gradient. Importantly, both checkpointed and non-checkpointed models exhibited the same spikes and dips in their validation losses over the course of training, also decreasing at similar rates.


\subsection{Extended analysis of cell state and update rule inductive biases}
\label{subsec:extended_analysis}

\begin{figure}[t]
  \centerline{\includegraphics[scale=1]{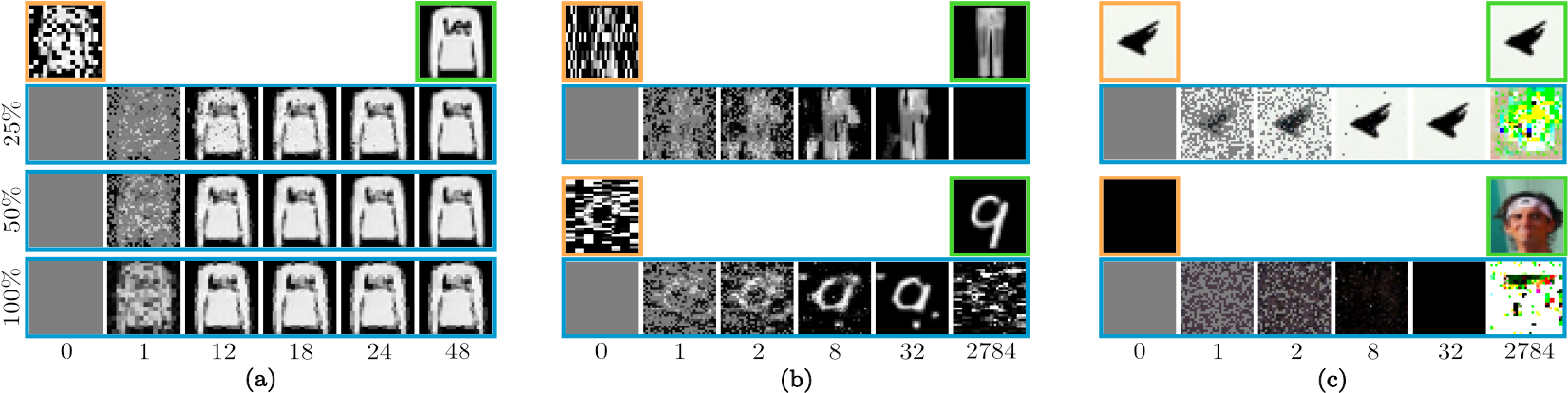}}
  \caption{Qualitative results showcasing UNetCA's inductive biases in terms of adapting to: \textbf{(a)} varying cell update rates; \textbf{(b)} noise configurations unseen during training, and; \textbf{(c)} unmasked and completely masked inputs. \textcolor{inGold}{Gold} boxes are inputs, \textcolor{gtGreen}{green} ground truths, and \textcolor{unetcaBlue}{blue} UNetCA outputs.\vspace{-\baselineskip}}
  \label{fig:unetca_extended_analysis}
\end{figure}

Here we present an extension of the analyses provided in \Sec{\ref{subsubsec:cell_state_analysis}} and \Sec{\ref{subsubsec:inductive_bias_investigation}}.

\paragraph{Adaptation to varying update rates (UNetCA).}
\Fig{\ref{fig:unetca_extended_analysis}} (a) shows UNetCA capable of adapting to a slower ($\sigma\!=\!25\%$) cell update rate despite being trained with a $\sigma\!=\!50\%$ cell update rate. Interestingly, UNetCA experiences difficulty synchronously updating all cells ($\sigma\!=\!100\%$), producing a noticeably lower quality output compared to its outputs at asynchronous rates. This is in contrast to ViTCA (\Fig{\ref{fig:analysis}} (d)), where the quality of output remains the same across all update rates. Also, not shown in \Fig{\ref{fig:analysis}} (d), but is important to note, are the number of ViTCA iterations from left-to-right, which are as follows: 1, 8, 12, 16, 32. We point attention to the fact that UNetCA required 48 iterations to converge with $\sigma\!=\!25\%$, 24 iterations to converge with $\sigma\!=\!50\%$, and could not converge to a good solution with $\sigma\!=\!100\%$, while ViTCA required 32 iterations to converge with $\sigma\!=\!25\%$, 16 iterations to converge with $\sigma\!=\!50\%$, and 8 iterations to converge with $\sigma\!=\!100\%$.

\paragraph{Generalization to noise unseen during training (UNetCA).}
As shown in \Fig{\ref{fig:unetca_extended_analysis}} (b), UNetCA is incapable of generalizing to noise configurations unseen during training, inducing a divergence in cell states. This is in contrast to ViTCA as shown in \Fig{\ref{fig:analysis}} (e). ViTCA not only produces a higher fidelity output mid-denoising, but it also maintains cell state stability.

\paragraph{Effects of not \vs completely masking input (UNetCA).}
\Fig{\ref{fig:unetca_extended_analysis}} (c; \emph{top}): Although UNetCA is able to successfully autoencode the unmasked input image, it eventually induces a divergence amongst cell states. This is in contrast to ViTCA as shown in \Fig{\ref{fig:analysis}} (g; \emph{left}). ViTCA not only produces a higher fidelity output mid-denoising, but it also maintains cell state stability. \Fig{\ref{fig:unetca_extended_analysis}} (c; \emph{bottom}): Unlike ViTCA (\Fig{\ref{fig:analysis}} (g; \emph{right})), UNetCA does not output the median image when attempting to denoise a completely masked input and instead causes cells to diverge.

\paragraph{Effect of masking heads.}
\begin{figure}[t]
  \centerline{\includegraphics[width=\textwidth]{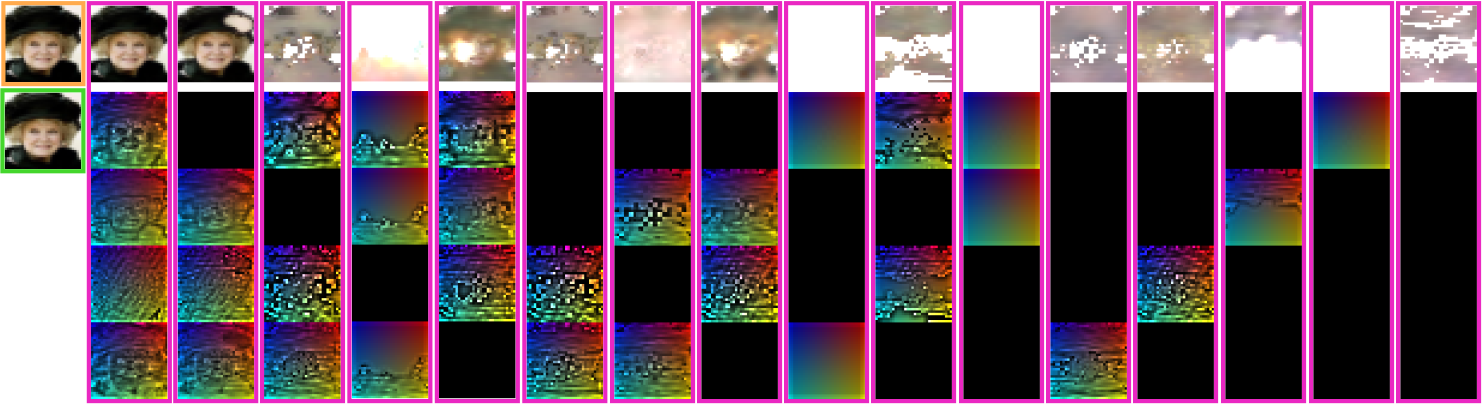}}
  \caption{Qualitative results showcasing ViTCA's inductive biases in terms of adapting to masking one or several of its self-attention heads. \textcolor{inGold}{Gold} boxes are inputs, \textcolor{gtGreen}{green} ground truths, and \textcolor{vitcaPurple}{purple} ViTCA outputs (after 2784 iterations). For reference, the first column of outputs does not contain any head masking.}
  \label{fig:attn_head_masking}
\end{figure}

\begin{wraptable}[15]{r}{0.515\textwidth}
  \caption{Profiling results showcasing ViTCA's runtime performance (forward and backward in milliseconds) and memory usage (in GB) while training on a minibatch of random $32 \times 3 \times H \times W$ images with gradient checkpointing disabled \vs\ enabled. We use $T\!=\!32$ ViTCA iterations and 16 checkpoint segments. Boldface denotes best results. Italicized items denote baseline configuration settings.\vspace{-0.5\baselineskip}}
  \label{tab:runtime_profiling_results} 
  \centering
  \small
  \setlength{\tabcolsep}{3pt}
    \begin{tabular}{@{}*{13}{l}@{}}
        & \diag{\fwdlabel} & \diag{\bwdlabel} & \diag{\memlabel} \\
        \toprule
        \textit{disabled} & \textbf{229ms} & \textbf{355ms} & 17.0GB \\
        enabled & 232ms & 576ms & \textbf{2.5GB} \\
        \bottomrule
    \end{tabular}
\end{wraptable}

\Fig{\ref{fig:attn_head_masking}} shows how ViTCA reacts to having its self-attention heads masked during autoencoding (no noise) an example from CelebA. The purpose of this experiment is to observe each head's contribution to the output. We can see that when none of the heads are masked, they attend to facial features and contours, and the output is as expected. However, once heads are masked, the unmasked heads stop attending to the features they once did and instead deteriorate. In some cases, the unmasked heads stop attending to anything at all. There are a couple of interesting cases: 1) When only the first head is masked, ViTCA is still able to successfully autoencode the input, although there is a slight degradation in quality. This is consistent with examples from the other datasets as well as when there is noise involved. 2) When certain heads are masked, the noise that the model was trained to denoise starts to appear (\eg fourth column from left and fifth column from right).

\subsection{Runtime analysis of ViTCA}
\label{subsec:runtime_analysis}

Here we provide a brief analysis of ViTCA's runtime performance and memory usage while training on a minibatch of random $32\times 3 \times H \times W$ images through measurements of forward pass duration (ms), backward pass duration (ms), and training memory usage (GB), with and without using gradient checkpointing. We use $T\!=\!32$ ViTCA iterations and 16 checkpoint segments. Results are shown in \Tab{\ref{tab:runtime_profiling_results}}. Gradient checkpointing provides substantial memory savings at the cost of proportionally increasing the duration of the backward pass.

\end{document}